\documentclass[letterpaper]{article} 

\usepackage{iclr2025_conference,times}
\usepackage{subcaption}
\usepackage{microtype}
\usepackage{url}
\usepackage{booktabs}

\usepackage{times}
\usepackage{longtable}
 \usepackage{latexsym}

\usepackage{amsmath}
\usepackage{adjustbox}
\usepackage[utf8]{inputenc} 
\usepackage[T1]{fontenc}    
\usepackage{booktabs}       
\usepackage{amsfonts}       
\usepackage{nicefrac}       
\usepackage{microtype}      
\usepackage{xcolor}         
\usepackage{multirow}
\usepackage{graphicx}
\usepackage{caption}
\usepackage{wrapfig}

\definecolor{applegreen}{rgb}{0.55, 0.71, 0.0}
\definecolor{canaryyellow}{rgb}{1.0, 0.94, 0.0}
\definecolor{daffodil}{rgb}{1.0, 1.0, 0.19}
\definecolor{amber}{rgb}{1.0, 0.75, 0.0}
\definecolor{aqua}{rgb}{0.0, 0.8, 1.0}
\definecolor{aureolin}{rgb}{0.9, 0.85, 0}
\usepackage{subcaption}
\usepackage{enumitem}
\definecolor{gainsboro}{rgb}{0.86, 0.86, 0.86}

\usepackage[textsize=tiny]{todonotes}

\title{\includegraphics[scale=0.04]{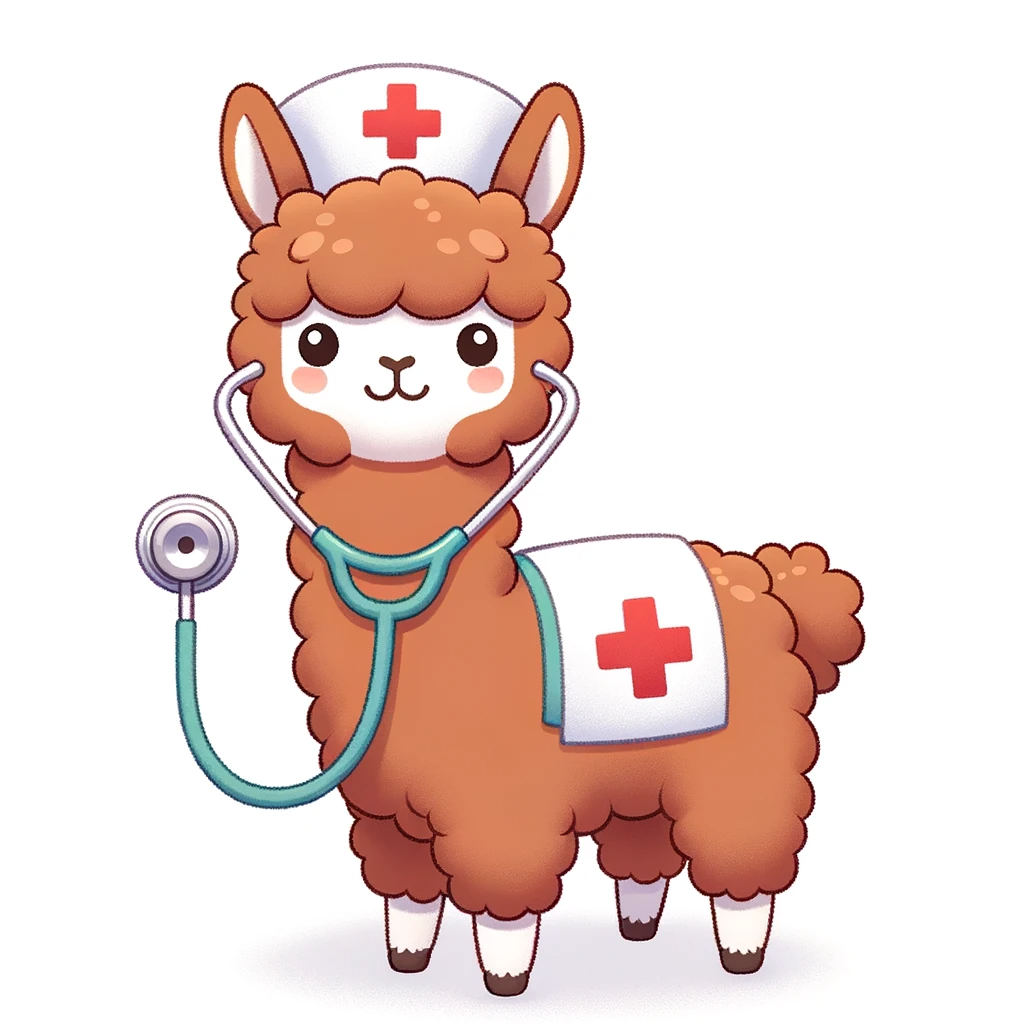}AlpaCare: Instruction Fine-tuned Large Language Models for Medical Applications}
\iclrfinalcopy
\begin{document}

\author{Xinlu Zhang$^1$\thanks{Corresponding Author: xinluzhang@ucsb.edu},  Chenxin Tian$^2$, Xianjun Yang$^1$, Lichang Chen$^3$, Zekun Li$^1$, Linda Ruth Petzold $^1$ \\
$^1$University of California, Santa Barbara \\
$^2$Chinese Academy of Medical Sciences and Peking Union                  
                        Medical College\\
$^3$ University of Maryland, College Park }

\newcommand{\fix}{\marginpar{FIX}}
\newcommand{\new}{\marginpar{NEW}}

\newcommand{\zekun}[1]{[\textcolor{cyan}{zekun: #1}]}


\maketitle

\begin{abstract}
Instruction-finetuning (IFT) has become crucial in aligning Large Language Models (LLMs) with diverse human needs and has shown great potential in medical applications. However, previous studies mainly fine-tune LLMs on biomedical datasets with limited diversity, which often rely on benchmarks or narrow task scopes, significantly limiting the effectiveness of their medical instruction-following ability and generalizability. To bridge this gap, we propose creating a diverse, machine-generated medical IFT dataset, \textit{MedInstruct-52k}, using GPT-4 and ChatGPT with a high-quality expert-curated seed set. We then fine-tune LLaMA-series models on the dataset to develop \textit{AlpaCare}. Despite using a smaller domain-specific dataset than previous medical LLMs, \textit{AlpaCare} not only demonstrates superior performance on medical applications, with up to 38.1\% absolute gain over best baselines in medical free-form instruction evaluations, but also achieves 6.7\% absolute gains averaged over multiple general domain benchmarks. Human evaluation further shows that \textit{AlpaCare} consistently outperforms best baselines in terms of both correctness and helpfulness. We offer data, model, and code publicly available \footnote{\url{https://github.com/XZhang97666/AlpaCare}}.

\end{abstract}

\section{Introduction}

Recent advancements in the training of large language models (LLMs) have placed a significant emphasis on instruction-finetuning (IFT), a critical step in enabling pre-trained LLMs to effectively follow instructions \citep{ouyang2022training, longpre2023flan, wei2022finetuned}. However, relying solely on NLP benchmarks to create instructional datasets can lead to `game-the-metric' issues, often failing to meet actual user needs \citep{ouyang2022training}. To better align with human intent, \citet{wang2023selfinstruct} introduces the concept of fine-tuning LLMs using diverse machine-generated instruction-response pairs. Subsequent works further highlight the importance of diversity in IFT datasets \citep{alpaca, xu2023wizardlm, vicuna2023,chen2023alpagasus}. However, how to improve dataset diversity in the medical domain for aligning with various user inquiries is still underexplored.

LLMs have demonstrated significant potential in the medical domain across various applications \citep{nori2023capabilities, nori2023generalist, singhal2022large, singhal2023expertlevel, liévin2023large, zhang2023enhancing}. To alleviate privacy concerns and manage costs, several medical open-source LLMs \citep{han2023medalpaca, li2023chatdoctor, wu2023pmcllama,xu2023baize} have been developed by tuning LLaMA \citep{ touvron2023llama,touvron2023llama-2} on medical datasets. Even substantial volumes, these datasets are limited in task scopes and instructions, primarily focusing on medical benchmarks or specific topics, due to the high cost of collecting real-world instruction datasets \citep{wang2023selfinstruct}, particularly when extending further into the medical domain\citep{jin2021disease,jin2019pubmedqa}. This lack of diversity can negatively impact the models' ability to follow instructions in medical applications and their effectiveness in the general domain. Therefore, there is an urgent need for a method to generate diverse medical IFT datasets that align with various domain-specific user inquiries while balancing cost.

\begin{wrapfigure}{r}{0.50\linewidth}
  \begin{center}
\includegraphics[width=\linewidth]{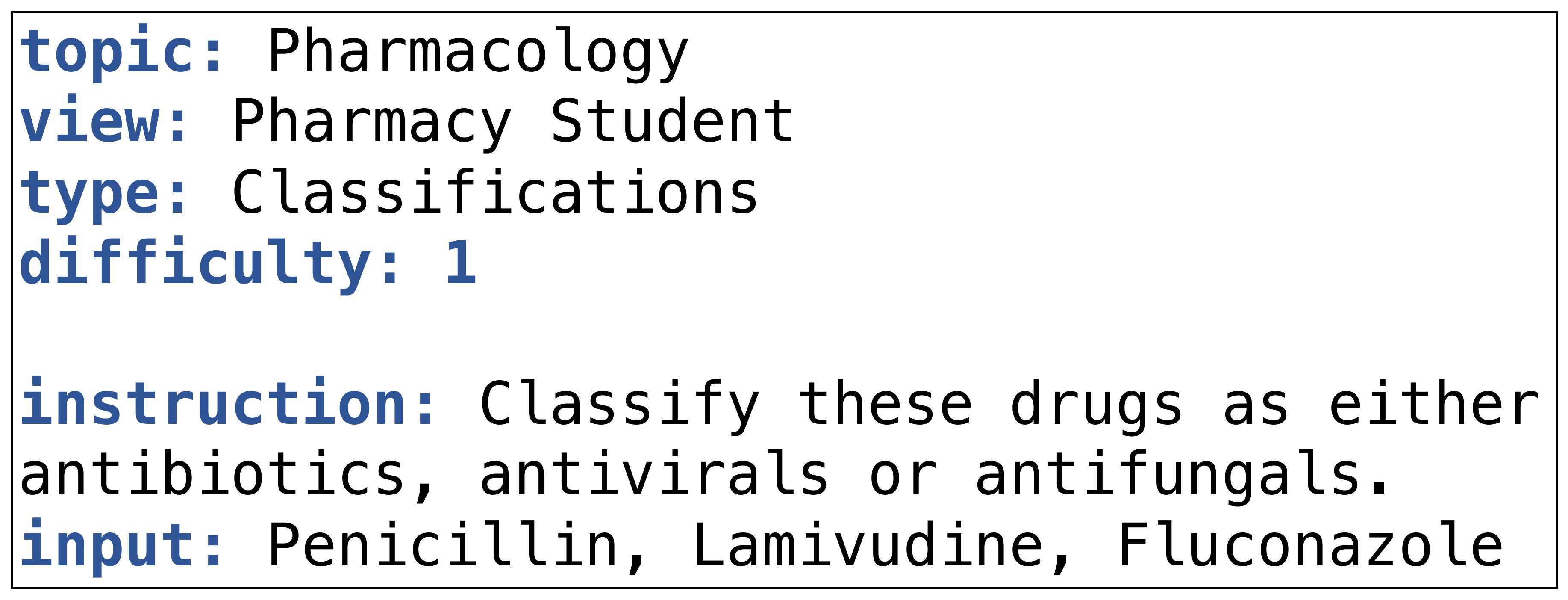}
\vspace{-8mm}
  \end{center}
\caption{\textbf{Selected example from the clinician-crafted seed set.} We focus on 4 perspectives: \textit{topic}, \textit{viewpoint}, \textit{task type}, and \textit{difficulty level}, to improve the seed set diversity. The set is further used to query GPT-4 to generate medical tasks.}
  \label{fig:seedexp}
    \vspace{-3mm}
\end{wrapfigure}

To bridge this gap, inspired by \citet{wang2023selfinstruct}, we propose a semi-automated process that uses GPT-4 \citep{openai2023gpt4} and ChatGPT \citep{openai2022chatgpt} to create a diverse medical IFT dataset for tuning a medical LLM, which can better align with various domain-specific user intents. Initially, to guide the overall task generations with meaningful medical instructions and considering different user needs, we create a high-quality seed set of 167 clinician-curated tasks spanning various medical topics, points of view, task types, and difficulty levels, as shown in Figure \ref{fig:seedexp}.

To automatically generate a broader array of tasks for training, we prompt GPT-4 to create instructions for new medical tasks by leveraging the existing clinician-curated tasks as demonstrations. After generating tasks and conducting deduplications, we employ ChatGPT to provide responses to the valid tasks. Consequently, we compile a 52k medical self-instruct dataset, \textit{MedInstruct-52k}, which supervises the tuning on the LLaMA series models \citep{touvron2023llama,touvron2023llama-2,llama3modelcard}, resulting in \textit{AlpaCare}. Due to the limited number of test sets available for evaluating medical LLMs in terms of instruction-following ability and medical capacity, we introduce a new clinician-crafted free-form instruction evaluation test set, \textit{MedInstruct-test}, covering medical tasks across different difficulty levels. 

Our comprehensive experiments within medical and general domains reveal that \textit{AlpaCare}, solely tuned on the 52k diverse medical IFT dataset, exhibits enhanced performance on medical applications and strong generalizability. It achieves up to a 38.1\% absolute gain over the best baselines in medical free-form instruction evaluations and a 6.7\% absolute gain averaged over multiple general domain benchmarks. Moreover, our human study on free-form instruction evaluations shows that \textit{AlpaCare} consistently produces better responses compared to existing medical LLMs by a large margin in terms of both correctness (+12\%) and helpfulness (+49\%).

Our paper makes the following contributions:
\begin{itemize}
    \item We propose a semi-automated, diverse medical IFT dataset generation pipeline to produce cost-effective, high-quality IFT data for LLM alignment in various medical applications.
    \item We conduct extensive experiments in medical and general domains, demonstrating that tuning LLMs with a diverse medical IFT dataset can boost their capacity in medical applications and generalization simultaneously.
    \item We release \textit{MedInstruct-52K}, a diverse machine-generated medical IFT dataset with 52K instruction-response pairs, and \textit{MedInstruct-test}, a test set of 216 clinician-crafted medical tasks, to help build and evaluate medical LLMs.

\end{itemize}

\section{Related Works}
\noindent\textbf{IFT.}
Closed-form IFT creates IFT datasets from existing NLP benchmarks using carefully designed instructions to improve model generalization on new tasks \citep{wei2022finetuned, sanh2022multitask, chung2022scaling, longpre2023flan}. However, these instructions are often simpler than real-world scenarios, leading to models that fail to align with diverse user intentions. In contrast, \citet{ouyang2022training} collects a diverse IFT dataset with real-world instructions and responses, rich in both instruction forms and task types. They train GPT-3 \cite{brown2020language} on this dataset to obtain InstructGPT \citep{ouyang2022training}, demonstrating promising results in aligning with diverse actual user needs. Due to the closed-source propriety of strong LLMs (e.g. ChatGPT and GPT-4), various open-source instruction fine-tuned models \citep{alpaca, xu2023wizardlm, vicuna2023, peng2023instruction} have been proposed to tune open-source LLMs using datasets obtained from these strong teacher models to enhance their instruction-following abilities. Alpaca \citep{alpaca} creates a 52k diverse machine-generated IFT dataset by distilling knowledge from the "teacher" Text-Davinci-003 \citep{hinton2015distilling,li2022explanations}. \citet{peng2023instruction} utilizes the same instructions with Alpaca but adopts GPT-4 as the "teacher" LLM to generate higher-quality and more diverse responses to improve the model's alignment on 3H (Helpfulness, Honesty, and Harmlessness) \citep{askell2021general}. Vicuna \citep{vicuna2023} is trained on the ShareGPT data \citep{sharegpt}, which contains actual ChatGPT users' diverse instructions, obtaining strong response quality and instruction-following ability. However, creating diverse IFT datasets for aligning models with various user intentions in the medical domain remains underexplored.

\noindent\textbf{LLMs in Biomedicine.}
Closed-source LLMs have demonstrated significant proficiency in the medical domain \citep{openai2022chatgpt,openai2023gpt4, singhal2023expertlevel,singhal2022large}. ChatGPT demonstrates promise in the US Medical Exam \citep{kung2023performance} and serves as a knowledge base for medical decision-making \citep{zhang2023enhancing}. The MedPaLM \citep{singhal2022large,singhal2023expertlevel} have shown performance in answering medical questions on par with that of medical professionals.
GPT-4 \citep{openai2023gpt4} obtains strong medical capacities without specialized training strategies in the medical domain or engineering for solving clinical tasks \cite{nori2023capabilities,nori2023generalist}. 
Due to privacy concerns and high costs, several open-source medical LLMs \citep{xu2023baize,han2023medalpaca, li2023chatdoctor, wu2023pmcllama} have been built
 by tuning open-source base models on medical corpus. ChatDoctor \citep{li2023chatdoctor} is fine-tuned using 100k online doctor-patient dialogues, while Baize-Healthcare \citep{xu2023baize} employs about 100k Quora and MedQuAD dialogues. MedAlpaca \citep{han2023medalpaca} utilizes a 230k dataset of question-answer pairs and dialogues. PMC-LLAMA \citep{wu2023pmcllama} continually trains LLaMA with millions of medical textbooks and papers, and then tunes it with a 202M-token dataset formed by benchmarks and dialogues during IFT. However, due to the high cost of collecting diverse real-world user instructions \citep{wang2023selfinstruct}, their datasets are limited in diversity, mainly focusing on medical benchmarks or within certain topics, such as doctor-patient conversations, hampering models' medical instruction-following ability and generalizability. We propose creating a cost-effective, diverse medical machine-generated IFT dataset using GPT-4 and ChatGPT to better align the model with various medical user intents. Other follow-up works after ours \citep{Xie2024MeLF,Tran2023BioInstructIT} consistently show the benefits of tuning medical LLMs with diverse machine-generated datasets.

\begin{figure*}[t!]
  \centering
\includegraphics[width=0.85\linewidth]{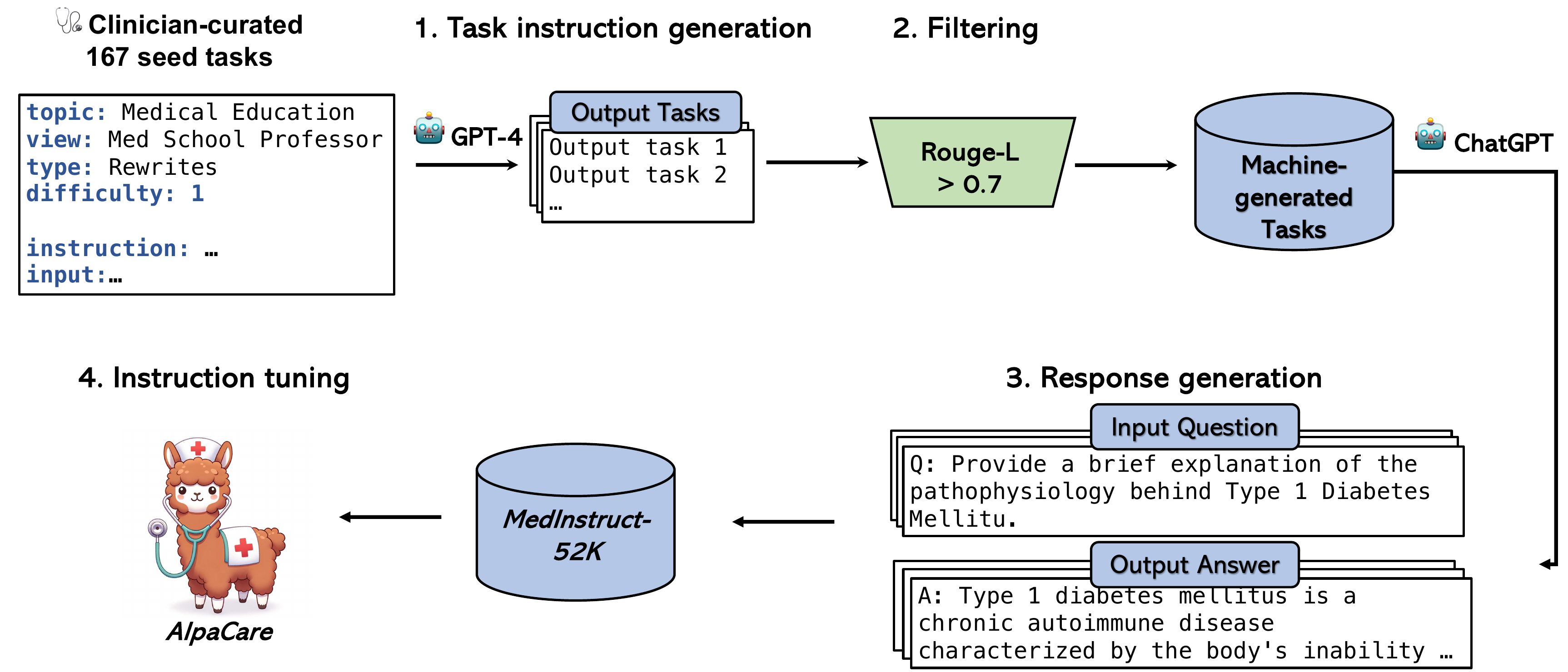}
\vspace{-3mm}
  \caption{\textbf{The pipeline of \textit{AlpaCare}.} The process starts with a small set of clinician-curated seed tasks. \textbf{1. Task instruction generation: }GPT-4 iteratively generates a series of new task instructions using 3 tasks from the seed set. \textbf{2. Filtering:} Ensures textual diversity by removing similar instructions via Rouge-L. \textbf{3. Response generation:} ChatGPT creates responses for each task, forming \textit{MedInstruct-52K}. \textbf{4. Instruction tuning:} The dataset is used to fine-tune LLaMA models, developing \textit{AlpaCare}.}
    \label{fig:pipeline}
      \vspace{-5mm}
\end{figure*}

\section{Method}

Collecting a large-scale medical IFT dataset is challenging because it necessitates 1) a deep understanding of the specific domain knowledge and 2) creativity in designing novel and diverse tasks by considering different real-world medical needs. To mitigate human effort while maintaining high quality, we propose a pipeline by instructing GPT-4 and ChatGPT to create a machine-generated dataset containing diverse domain-specific tasks. The process starts with utilizing a small set of high-quality clinician-curated seed tasks with 167 instances to prompt GPT-4 in generating medical tasks. Similar instructions are removed from the generated medical tasks, preserving 52k instances which are subsequently inputted into ChatGPT for response generation. The instruction-response pairs dataset, \textit{MedInstruct-52k}, is used to tune the LLaMA, resulting in \textit{AlpaCare} with superior medical instruction-following ability and generalizability. The pipeline is shown in Figure \ref{fig:pipeline}.

\vspace{-2mm}
\subsection{Clinician-curated Seed Dataset} \label{sec:seed}

A diverse and high-quality seed task is essential for prompting LLMs in task generation \citep{wang2023selfinstruct}. We focus on 4 key areas, taking into account various user intents in medical applications, to improve the diversity of seed instructions: \textit{topic}, \textit{view}, \textit{type}, and \textit{difficulty level}. Specifically, the \textit{topic} covers various submedical domains, such as radiology, genetics, and psychophysiology. The \textit{view} is derived from diverse medical personnel, including researchers, medical students, and patients, who have different inquiries, to ensure a comprehensive range of viewpoints based on various levels of domain knowledge. For the \textit{type}, we include various task formats, such as summarization, rewriting, single-hop, and multi-hop reasoning, to align with different application needs. Lastly, each task is categorized by its medical \textit{difficulty level}, ranging from 1 to 5 (low to high), to ensure that the seed tasks can prompt new tasks on a wide range of expertise levels. We defer the explanation of the difficulty score to Appendix \ref{app:difficulty} for further clarification. A clinician crafts each task considering these 4 dimensions, and each task contains instruction and may have a corresponding input, which could be a detailed medical example to further elucidate the instruction and enhance task diversity. Examples are shown in Figure \ref{fig:seedexp}.

\vspace{-2mm}
\subsection{Medical IFT Dataset Generation and LLM Tuning}
\vspace{-2mm}
We utilize GPT-4 for in-context learning by randomly selecting 3 tasks from the seed set and generating 12 tasks for each run. To ensure generated task diversity, we instruct GPT-4 to consider the 4 aspects outlined in \ref{sec:seed}. To further amplify textual diversity, instructions with a Rouge-L similarity above 0.7 to any other generated task are discarded \citep{wang2023selfinstruct}. Due to the lengthy propriety of medical text, we separately generate responses for each task using ChatGPT (GPT-3.5-turbo), which has demonstrated efficacy in the medical domain \citep{zhang2023enhancing}. Finally, we result in 52k machine-generated medical instruction-response pairs, \textit{MedInstruct-52k}. Detailed prompts for instruction and output generation are provided in the Appendix \ref{app:generation_prompt}. To verify \textit{medInstcut-52k}'s data quality, we randomly select 50 instances for a clinician to evaluate, resulting in 49 out of 50 responses being graded as correct, which demonstrates the dataset's high quality. The LLM APIs cost analysis and chosen reason are in Appendix \ref{app:cost}. 

During IFT, we adopt the same training prompt and hyper-parameter setup as \citet{alpaca} to fine-tune LLaMA models on \textit{MedInstruct-52k}, Specifically, we employ instructions and inputs (when available) as inputs to tune the model to generate corresponding response outputs through a standard supervised fine-tuning with cross-entropy loss. We defer hyper-parameter setup into Appendix \ref{app:hyper}.

\begin{table*}[t!]
\centering
  \caption{\textbf{Comparative analysis of free-form instruction evaluation.} Performance comparison of \textit{AlpaCare} and instruction-tuned baselines. GPT-3.5-turbo acts as a judge for pairwise auto-evaluation. Each instruction-tuned model is compared with 4 distinct reference models: Text-davinci-003, GPT-3.5-turbo, GPT-4, and Claude-2. `AVG' denotes the average performance score across all referenced models in each test set.}
    \vspace{-3mm}
  \resizebox{\linewidth}{!}{
      \Huge
    \begin{tabular}{l ccccc ccccc}
     \toprule
     & \multicolumn{5}{c}{\textbf{iCliniq}} & \multicolumn{5}{c}{\textbf{MedInstruct}} \\
     \cmidrule(lr){2-6}\cmidrule(lr){7-11} 
     & Text-davinci-003 & GPT-3.5-turbo & GPT-4& Claude-2 & AVG &  Text-davinci-003 & GPT-3.5-turbo & GPT-4& Claude-2 & AVG \\
  \midrule
    Alpaca & 38.8 & 30.4 & 12.8 & 15.6 & 24.4 & 25.0& 20.6 & 21.5 & 15.6 & 22.5\\
    ChatDoctor & 25.4 & 16.7 & 6.5 & 9.3 & 14.5 & 35.6 & 18.3 & 20.4 & 13.4 & 18.2 \\
    Medalpaca & 35.6 & 24.3 & 10.1 & 13.2 & 20.8 & 45.1& 33.5 & 34.0 & 29.2 &  28.1\\
    PMC & 8.3 & 7.2 & 6.5 & 0.2 &5.5 & 5.1 & 4.5 & 4.6 & 0.2 &  4.6 \\
    Baize-H & 41.8 & 36.3 & 19.2 & 20.6 &29.5 & 35.1 & 22.2 & 22.2 & 15.6 & 26.6\\
    AlpaCare & \textbf{66.6} & \textbf{50.6} & \textbf{47.4} & \textbf{49.7} &   \textbf{53.6}& \textbf{67.6}& \textbf{49.8} & \textbf{48.1} & \textbf{48.4} & \textbf{53.5} \\
     \bottomrule  
    \end{tabular}
  }
 \label{tab:med-free-form}
\end{table*}

\begin{table}[t!]
\centering
  \caption{\textbf{Results on medical benchmarks.} `AVG' represents the mean performance score across tasks.}
    \vspace{-3mm}
  \resizebox{0.85\linewidth}{!}{
 \Huge
    \begin{tabular}{lcccccc}
     \toprule
      & MEDQA & HeadQA & PubmedQA & MEDMCQA & MeQSum  & AVG\\
  \midrule
    Alpaca & 35.7 & 29.1 & \textbf{75.4} & 29.2 & 24.4 & 38.8\\
    ChatDoctor & 34.3 & 30.0 & 73.6 & \textbf{33.5} & 27.1 &  39.7\\
    Medalpaca & \textbf{38.4} & 30.3 & 72.8 & 31.3 & 11.0 & 36.8\\
    PMC & 34.2 & 28.1 & 68.2 & 26.1 & 9.2 & 33.2\\
    Baize-H & 34.5 & 29.3 & 73.8 & 32.5 & 8.1 & 35.6 \\
    AlpaCare & 35.5 & \textbf{30.4} & 74.8 &\textbf{33.5} & \textbf{29.0} & \textbf{40.6}\\
     \bottomrule  
    \end{tabular}
  }
 \label{tab:benchmark}
\end{table}

\section{Experimental Setup}
\subsection{Free-form Instruction Evaluation}  
\vspace{-1mm}
\noindent\textbf{Datasets.}To evaluate the effectiveness of LLMs in a user-oriented manner, we conduct free-form instruction evaluations on two medical datasets. (1) iCliniq\footnote{We randomly selected 1,000 instances for evaluation from the 10,000 instances proposed by \citet{li2023chatdoctor}.}, a dataset comprising transcripts of real patient-doctor conversations collected from an online website \citep{li2023chatdoctor}. In this task, the model processes patient inquiries as input and then simulates a doctor to provide corresponding answers. (2) \textit{MedInstruct-test}, a dataset created by our clinicians, includes 216 medical instructions. These instructions mimic inquiries posed by different medical personnel, varying in difficulty on a scale from 1 to 5, with 1 being the simplest and 5 being the most challenging. We defer the difficulty levels description and test set statistics into the Appendix \ref{app:difficulty} and \ref{app:test_stat}, respectively.

\noindent\textbf{Evaluation Metric.} \label{sec:eval}
We conduct auto-evaluation by employing GPT-3.5-turbo to serve as a judge \citep{zheng2023judging}. The judge pairwise compares responses from a model with reference responses produced by another LLM API for each instruction in the test sets. To conduct a holistic evaluation, we employ reference outputs generated by 4 different APIs: Text-davinci-003, GPT-3.5-turbo, GPT-4 and Claude-2, respectively.
To ensure unbiased evaluation and avoid positional bias \citep{wang2023large}, we evaluate each output comparison twice, alternating the order of the model output and the reference output. We follow \citet{alpaca_eval} to score our models and baselines by calculating the win rate. To ensure fair comparisons, we set the maximum token length to 1024 and utilize greedy decoding for the generation of all model outputs and reference responses.

\vspace{-2mm}
\subsection{Benchmark Evaluation}
\vspace{-1mm}
\noindent\textbf{Datasets.} We further evaluate \textit{AlpaCare} on 4 medical multiple-choice benchmarks,
namely MedQA \citep{jin2021disease}, HeadQA \citep{headqa}, PubmedQA \citep{jin2019pubmedqa}, and MedMCQA \citep{pal2022medmcqa},
as well as a summarization dataset, i.e., MeQSum \citep{MedQsum}\footnote{We randomly selected 200 out of 1000 instances in MeQSum.},
to assess the model’s medical capacity.

\noindent\textbf{Evaluation Metric.}
Following \citet{eval-harness}, we conduct the multiple-choice benchmark evaluation and report the accuracy. For the summarization task, we utilize greedy decoding with a maximum token length of 1024 to generate outputs and report the ROUGE-L score.

\vspace{-2mm}
\subsection{Baselines}
\vspace{-1mm}
We evaluate the performance of \textit{AlpaCare} by comparing it with both general and medical LLMs based on the LLaMA models. We consider a range of models including: (1) Alpaca, tuning on 52k general domain machine-generated samples with responses from Text-davinci-003; (2) ChatDoctor, fine-tuning with 100k real patient-doctor dialogues; (3) MedAlpaca, utilizing approximately 230k medical instances such as Q\&A pairs and doctor-patient conversations; (4) PMC-LLaMA (PMC), a two-step tuning model that was first trained on 4.8 million biomedical papers and 30k medical textbooks, then instruction-tuned on a corpus of 202 million tokens; and (5) Baize-Healthcare (Baize-H), training with around 100k multi-turn medical dialogues.

\vspace{-2mm}
\section{Experiment Results}
\subsection{Main Results}
\vspace{-1mm}
\noindent\textbf{Free-form Instruction Evaluation Performance.} The evaluation results for 4 reference models on both datasets are summarized in Table \ref{tab:med-free-form}. \textit{AlpaCare} outperforms its general domain counterpart, Alpaca, demonstrating that domain-specific training bolsters medical capabilities. Despite tuning with only 52k medical instruction-response pairs, \textit{AlpaCare} consistently and significantly surpasses other medical models, which are trained on considerably larger datasets, across various reference LLMs. Specifically, for average scores across reference models, \textit{AlpaCare} demonstrates a relative gain of 130\% on iCliniq and 90\% on MedInstruct, respectively, compared to the best baselines. These results highlight the advantages of improving medical proficiency by training with a diverse, domain-specific IFT dataset. Surprisingly, medical LLMs don't always outperform general ones in medical tasks, and some even fail to generate useful responses, possibly due to their limited training scope restricting conversational skills.

\noindent\textbf{Benchmark Evaluation Performance.}
Table \ref{tab:benchmark} presents an extensive evaluation of \textit{AlpaCare} on 5 medical benchmarks. \textit{AlpaCare} obtain the best performance on average, highlighting its robust capability in the medical domain. Benchmarks evaluate a model's intrinsic knowledge\citep{eval-harness}, which is mainly gained in LLM pretraining instead of instruction fine-tuning \citep{zhou2023lima}.  
\textit{AlpaCare}'s strong medical capability, combined with its superior ability to follow medical instructions, enables it to meet a wide range of medical application needs effectively.

\begin{table}[t!]
\centering
  \caption{\textbf{Performance on general domain tasks.} AlpacaFarm is a free-form instruction evaluation, MMLU and BBH are knowledge benchmarks and TruthfulQA is a truthfulness task. `AVG' denotes the average score across all tasks.}  
    \vspace{-3mm}
  \resizebox{0.65\linewidth}{!}{
    \begin{tabular}{lccccc }
     \toprule
      & AlpacaFarm  & MMLU & BBH & TruthfulQA & AVG \\
  \midrule
    Alpaca &  22.7 & 40.8 & 32.4 & 25.6 & 30.4\\
    ChatDoctor & 21.2 & 34.3 & 31.9 &  \textbf{27.8} & 28.8 \\
    Medalpaca & 25.8 & 41.7 & 30.6 & 24.6 & 30.7 \\ 
    PMC & 8.3 & 23.6 & 30.8 &  23.8 & 21.6 \\
    Baize-H & 18.3 & 36.5 & 30.1 & 23.5 & 27.1 \\
    AlpaCare & \textbf{40.7} &  \textbf{45.6} & \textbf{34.0} & 27.5 & \textbf{37.0}\\
     \bottomrule  
    \end{tabular}
  }
      \vspace{-3mm}
 \label{tab:generalizability}
\end{table}

\vspace{-2mm}
\subsection{Generalizability Evaluation}
\vspace{-1mm}
Training models with specific data may lead to catastrophic forgetting, limiting their generalizability  \citep{kirkpatrick2017overcoming}. Our approach, instruction tuning a model with a diverse, domain-specific dataset, aims to improve its generalizability simultaneously. We test this using \textit{AlpaCare} in AlpaFarm \citep{dubois2023alpacafarm}, MMLU \citep{hendrycks2021measuring}, BBH \citep{suzgun2022challenging} and TruthfulQA \citep{lin2022truthfulqa}. We compare \textit{AlpaCare} with 4 reference LLMs in AlpaFarm and report the average score, and follow \cite{chia2023instructeval} to holistically evaluate models' general domain knowledge on MMLU (5-shot) and BBH (3-shot), receptively; and evaluate the model truthfulness on TruthfulQA (0-shot) with \citet{eval-harness}. The results are shown in Table \ref{tab:generalizability}. The detailed score for each reference model on AlpaFarm and more general domain experiment are deferred to Table \ref{tab:gen-free-form_all} and Table \ref{app:moregeneral} in the Appendix \ref{app:moreexp}.

Medical LLMs often have worse or comparable results than the general LLM, Alpaca, in terms of generalizability. However, AlpaCare significantly outperforms both medical and general domain baselines in multiple general tasks on average. Specifically, \textit{AlpaCare} shows a significant relative improvement of 57.8\% on AlpacaFarm compared to the best baseline, demonstrating strong general instruction-following ability. Moreover, \textit{AlpaCare} scores higher in general knowledge tasks and maintains comparable truthfulness scores compared to other baselines, indicating strong generalization abilities due to high data diversity.

\subsection{Ablation Study}
\vspace{-1mm}
 To further understand the effectiveness of \textit{AlpaCare}, we conduct systematic ablation studies on two medical free-form instruction evaluations and report mean results of each task across 4 reference models, receptively. We defer detailed results of each reference model into Appendix \ref{app:abstudy}.

\noindent\textbf{\textit{AlpaCare} consistently delivers superior performance in 13B model comparisons.}
To explore the impact of scaling up LLM size, we fine-tune \textit{AlpaCare}-13B on LLaMA-13B and compare its performance against other 13B LLMs. Results are shown in Table \ref{tab:13bmodel}.

\textit{AlpaCare}-13B consistently outperforms other 13B models in both tasks. This reaffirms the conclusion drawn from the 7B model comparison: tuning models with a diverse medical IFT dataset can better align the model with user needs across different medical applications.

\begin{table}[t!]
  \begin{minipage}[b]{0.49\linewidth} \centering
  \caption{ \textbf{Result comparison on 13B instruction-tuned models.}}
    \vspace{-3mm}
    \Huge
  \resizebox{0.95\linewidth}{!}{
    \begin{tabular}{lcccc}
     \toprule
      & Alpaca  & Medalpaca & PMC &  AlpaCare \\
  \midrule
  iCliniq & 31.3 & 3.9 & 25.4& \textbf{54.4} \\
  MedInstruct & 26.9& 0.1& 34.7& \textbf{54.5} \\
     \bottomrule  
    \end{tabular}
  }
 \label{tab:13bmodel}
  \end{minipage}
  \hfill 
  \begin{minipage}[b]{0.49\linewidth} \centering
 \caption{\textbf{Results on different LLM backbones.}}
      \vspace{-3mm}
  \resizebox{0.95\linewidth}{!}{%
   \Huge
	\begin{tabular}{llccc}
	 \toprule
	  & & LLaMA & LLaMA-2 & LLaMA-3 \\
  	 \midrule
	 \multirow{2}{*}{iCliniq} & Alpaca & 24.4 & 30.3 & 26.8 \\
	  & AlpaCare & \textbf{53.6} & \textbf{53.7} & \textbf{56.9} \\
  	 \midrule
	 \multirow{2}{*}{MedInstruct} & Alpaca & 23.2 & 26.8 & 20.7 \\
	  & AlpaCare & \textbf{53.5} & \textbf{54.2} & \textbf{56.6} \\
	 \bottomrule
	\end{tabular}
  }
 \label{tab:model_type}
  \end{minipage}
\end{table}

\noindent\textbf{\textit{AlpaCare} achieves superior performance across various backbones.} To explore the effect of different LLM backbones, we tune Alpaca-LLaMA2/3 and \textit{AlpaCare}-LLaMA2/3 by training LLaMA2-7B \citep{touvron2023llama-2}  and LLaMA3-8B \citep{llama3modelcard} on Alpaca data and \textit{MedInstruct-52k}, respectively.  Table \ref{tab:model_type} compares the performance of Alpaca and  \textit{AlpaCare} based on different LLM backbone families.

Consistent with the results of using LLaMA-1 as the backbone, \textit{AlpaCare}-LLaMA2/3 consistently and significantly outperforms Alpaca-LLaMA2/3 in both datasets. This further underscores the backbone agnostic property of our method and emphasises tuning with a diverse medical IFT dataset can bolsters models' medical capabilities.


\begin{wraptable}{r}{0.4\textwidth}
\centering
  \caption{
   \textbf{Results evaluated by the different judge.} Free-form instruction evaluation with Claude-2 as the judge.
  }
    \vspace{-3mm}
  \resizebox{0.95\linewidth}{!}{
    \begin{tabular}{lcc}
     \toprule
      & iCliniq & MedInstruct \\
  \midrule
  Alpaca & 26.7 & 23.5 \\
  ChatDoctor & 17.4 & 21.7 \\
  Medalpaca & 26.7 & 23.1 \\
  PMC & 1.3 & 1.8 \\
  Baize-H & 25.5 & 19.8 \\
  AlpaCare & \textbf{38.8} & \textbf{31.5} \\
     \bottomrule  
    \end{tabular}  }
\label{tab:judge_type}
\end{wraptable}

\noindent\textbf{\textit{AlpaCare} shows robust performance across different judges.}
Recent studies have highlighted potential biases in the LLM evaluator \citep{wang2023large}. ChatGPT may give a higher preference for outputs from ChatGPT and GPT-4, which are both trained by OpenAI. To robustly evaluate our method, we introduce an alternative judge, Claude-2 \citep{claude-2} from Anthropic, to mitigate the potential biases of relying on a single family of judges. The results are shown in Table \ref{tab:judge_type}.

Upon evaluation by Claude-2, it is observed that \textit{AlpaCare} consistently outperforms its IFT baselines by a large margin. This aligns with findings from assessments using GPT-3.5-turbo as the judge. Such consistency underscores the superior medical proficiency of our approach.

\section{Human Study}\label{sec:humanstudy}

\begin{wrapfigure}{r}{0.55\linewidth}
  \centering
  \vspace{-3mm}
\includegraphics[width=0.85\linewidth]{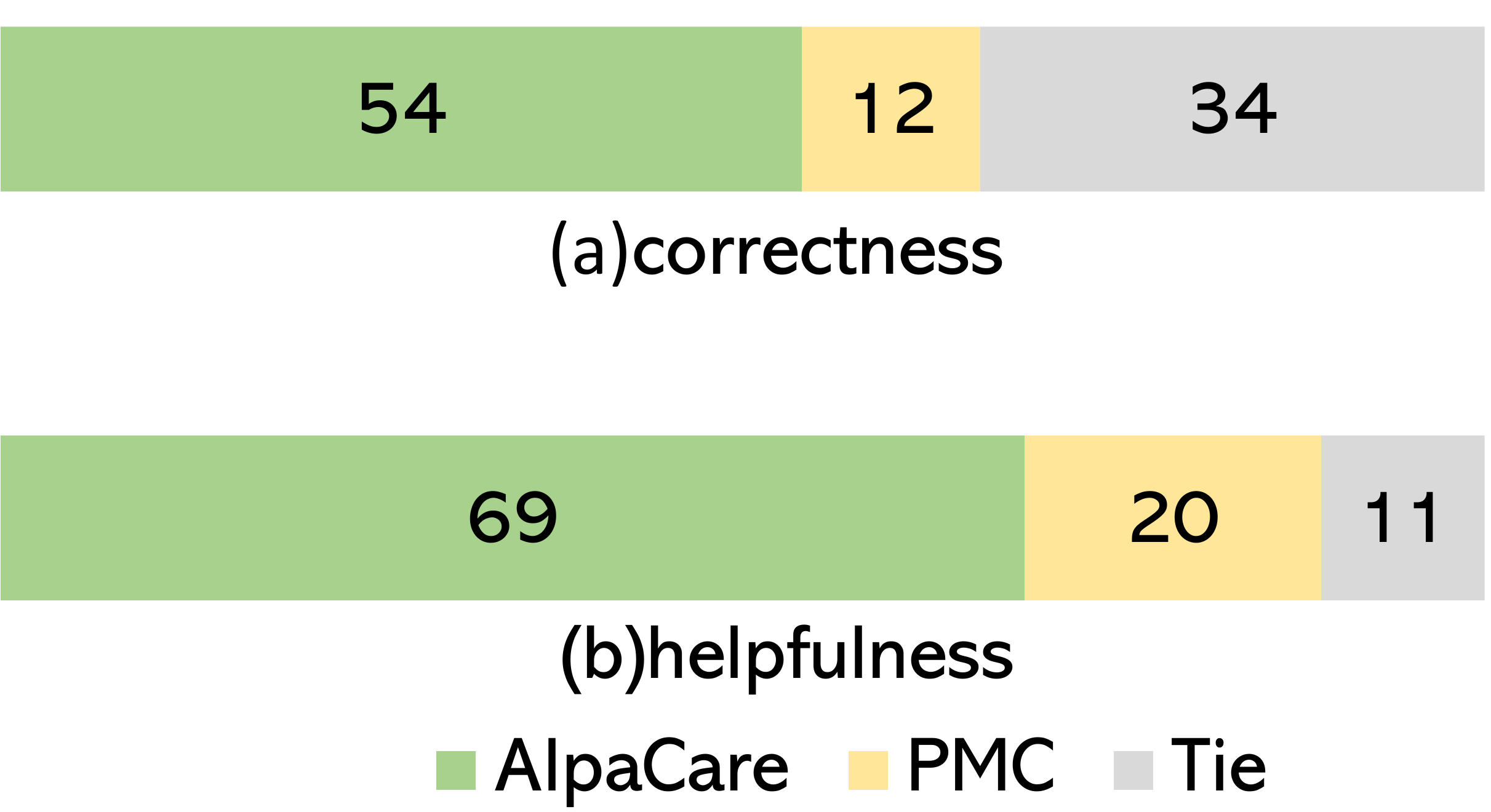}
  \vspace{-3mm}
  \caption{\textbf{Human study results.} Head-to-head clinician preference
comparison between \textit{AlpaCare}-13B and PMC-13B on (a) correctness and (b) helpfulness.}
      \label{fig:humaneval}
\vspace{-3mm}
\end{wrapfigure}

We further conduct human studies to label question-and-answer pairs in medical free-form instruction evaluation. Three annotators with MD degree in progress are involved in the study to perform pairwise comparisons for each question and answer pair. Specifically, we randomly select 50 prompts from each test set, totaling 100 prompts. These prompts, along with the responses generated by both \textit{AlpaCare}-13B and PMC-13B, the best baseline in the 13B models, are presented to the annotators for evaluation. The evaluation is based on two criteria: correctness and helpfulness. Correctness evaluates whether the response provides accurate medical knowledge to address question posed, while helpfulness measures the model's ability to assist users concisely and efficiently, considering the user intent. In practical terms, an answer can be correct but not necessarily helpful if it is too verbose and lacks guidance. To determine the final result for each criterion of each evaluation instance, we employ a majority vote method. If at least two of the annotators share the same opinion, their preference is considered the final answer; otherwise, we consider the outputs of the two models to be tied. The results are shown in Figure \ref{fig:humaneval}.

Consistent with previous results, \textit{AlpaCare}-13B outperforms PMC-13B in human evaluation, with 54\% of answers preferred by expert annotators for correctness and 69\% for helpfulness. This demonstrates \textit{AlpaCare}'s superior medical capacity and practical usability.  The greater improvement in helpfulness over correctness for \textit{AlpaCare} is expected, as the goal of IFT is to enhance LLMs' instruction-following ability to meet diverse user needs, rather than acquiring new knowledge.

\vspace{-3mm}
\section{Analysis \& Case Study}
\vspace{-3mm}
\subsection{IFT Dataset Diversity Analysis}

\vspace{-3mm}

\begin{figure*}[t!]
  \centering
  \vspace{-3mm}
\includegraphics[width=0.9\linewidth,height=0.45\linewidth]{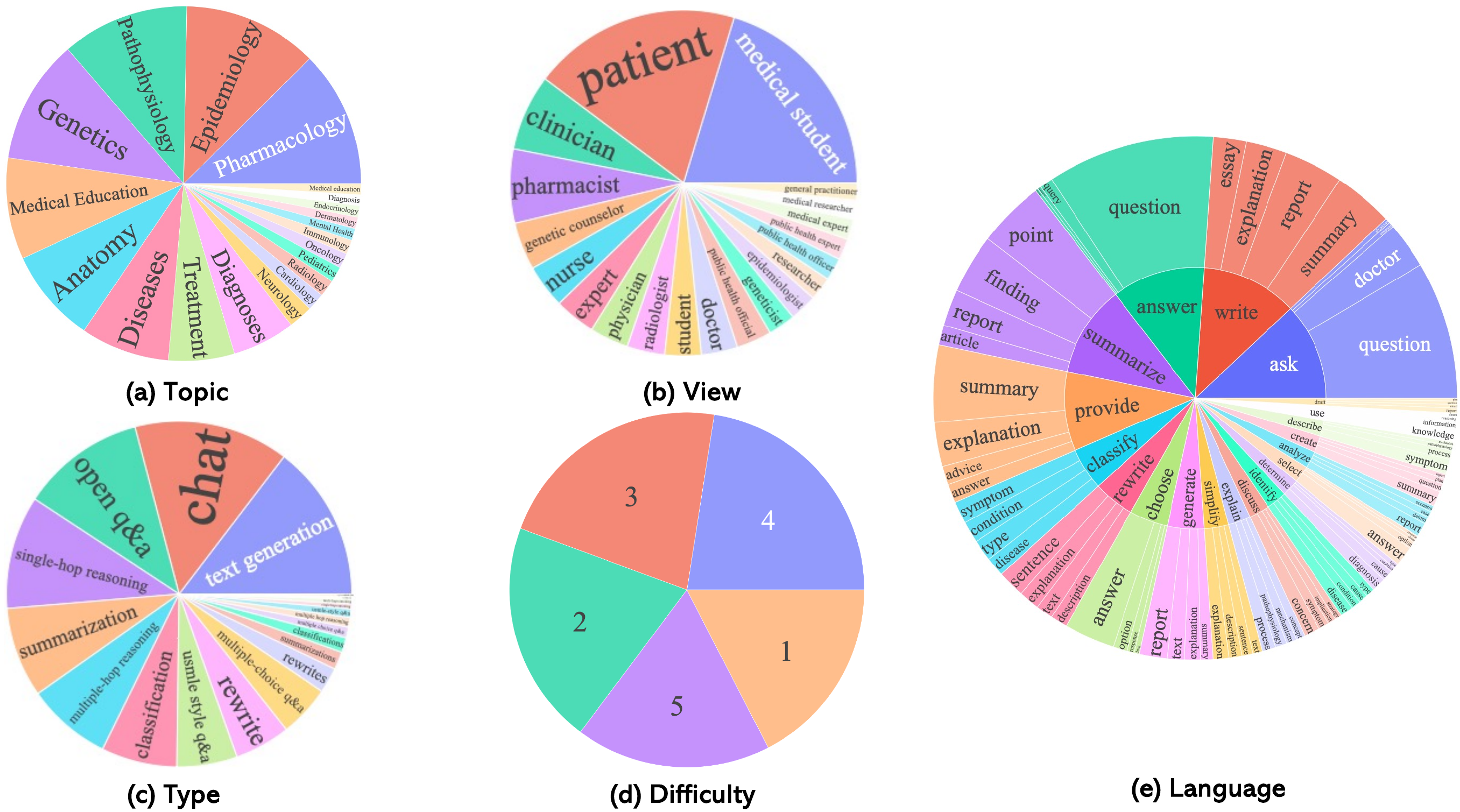}
  \caption{\textbf{Analysis of diversity in the \textit{MedInstruct-52k}.} In panels (a-c), the top 20 entries for topic, view, and type are displayed, respectively. Panel (d) shows the distribution of instruction medical difficulty levels. Panel (e) analyzes linguistic diversity to depict the top 20 root verbs in the inner circle and their 4 primary direct noun objects in the outer circle in the generated instructions.}
    \label{fig:data_stat}
          \vspace{-7mm}
\end{figure*}

Training a model with diverse instructions enhances its ability to follow instructions \citep{wang2023selfinstruct}.
However, current medical LLMs often have training data lacking in instructional diversity, typically using repetitive instructions across different instances \citep{li2023chatdoctor,han2023medalpaca,wu2023pmcllama}. 
To examine the diversity in our dataset, we plot the distributions of 4 key areas for instruction generation from \textit{MedInstruct-52k}, shown in Figure \ref{fig:data_stat} (a)- (d). Specifically, we present the top 20 topics, views and types and the difficulty levels from 1 to 5, offering insight into training data distribution. We further analyze instruction linguistic diversity by showing the root verbs and their corresponding direct-object nouns from each instruction. The top 20 root verbs and their 4 most common direct-object nouns are displayed in Figure \ref{fig:data_stat} (e), representing 22\% of the total dataset. 
Our findings show quite diverse medical intents and textual formats in our \textit{MedInstruct-52k}.

\begin{wraptable}{r}{0.6\textwidth}
\centering
\vspace{-2mm}
  \caption{\textbf{Quantitative comparison of linguistic diversity in medical instructional datasets.} Comparing linguistic entropy of each IFT dataset for medical LLMs. The higher value represents better diversity.}
      \vspace{-3mm}
\resizebox{\linewidth}{!}{%
   \Huge
	\begin{tabular}{l c c c c c  }
	 \toprule
	  &  ChatDoctor &	Medalpaca & PMC	& Baize-H& AlpaCare\\
  	 \midrule
    Entropy &0& 0&2.85&3.45 &\textbf{5.57}\\
	 \bottomrule
	\end{tabular}
  }
\vspace{-3mm}
  \label{tab:entropy}
  \end{wraptable}
  
To quantitatively showcase our dataset's diversity in comparison to IFT dataset of other medical LLMs, we calculate the linguistic entropy in the instructions of instruction-following datasets used for medical models. Higher entropy values signify greater diversity. Specifically, we analyze the top 20 root verbs and their 4 primary direct noun objects for each dataset and calculate verb-noun pair entropy, as shown in Table \ref{tab:entropy}. \textit{AlpaCare}'s dataset, \textit{MedInstruct-52K}, exhibits the highest entropy, underscoring its superior diversity, which in turn better elicits the model's instruction-following capabilities during IFT.

\subsection{Generation Case Study}

\begin{figure*}[t!]
  \centering
\includegraphics[width=\linewidth,height=0.85\linewidth]{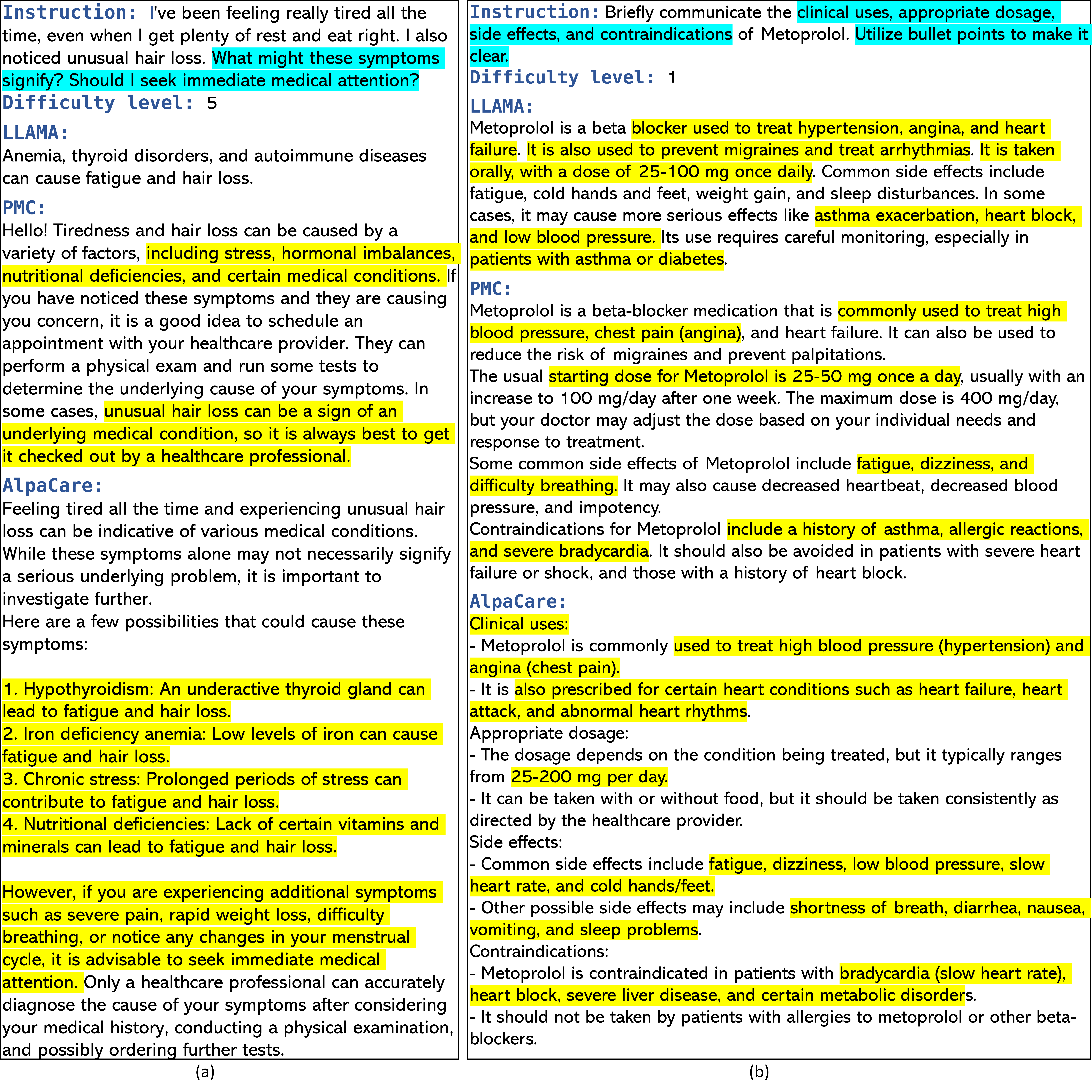}
  \vspace{-4mm}
\caption{\textbf{Case Study} of 13B models of \textit{AlpaCare} and PMC on (a) correctness and (b) helpfulness. Instruction key points and primary responses are highlighted in {\color{aqua}blue} and {\color{aureolin}yellow}, respectively.}
    \label{fig:casestudy}
      \vspace{-7mm}
\end{figure*}

We randomly selected one win case from \textit{MedInstruct-test} for correctness and another for helpfulness, as described in Section \ref{sec:humanstudy}. Figure \ref{fig:casestudy} displays the instructions and outputs of the base model, LLAMA-13B, and 13B medical models, \textit{AlpaCare} and PMC.

Figure \ref{fig:casestudy}(a) illustrates a correctness case 
with a high medical difficulty level, where LLAMA-13B struggles to provide accurate responses, while Medical LLMs show improvement. However, PMC provides a general overview, mentioning common causes like stress and hormonal imbalances, and uses vague terms such as 'certain medical conditions,' which lack specificity and do not provide actionable medical insights. In contrast, \textit{AlpaCare} gives a detailed analysis, identifying specific conditions like hypothyroidism and iron deficiency anemia, and emphasizes the importance of medical attention for severe symptoms, enhancing the guidance's precision and actionability.

Figure \ref{fig:casestudy}(b) shows a helpfulness case study. Both the base and medical models accurately describe Metoprolol, but LLAMA and PMC disregard the requested bullet-point format, reducing clarity. In contrast, \textit{AlpaCare} effectively follows the instructions with well-organized formatting. This demonstrates that fine-tuning with a diverse medical IFT dataset enhances the model's ability to follow instructions, thereby increasing helpfulness.

\vspace{-3mm}
\section{Conclusion}
\vspace{-3mm}
In this paper, we propose a semi-automated pipeline using GPT-4 and ChatGPT to create the diverse \textit{MedInstruct-52k} dataset for LLM tuning. Extensive experiments on multiple benchmarks show that when trained on this dataset, our model, \textit{AlpaCare}, demonstrates stronger medical and general instruction-following capabilities compared to medical LLM baselines, underscoring the importance of data diversity in medical AI model development.


\clearpage
\newpage
\bibliography{ref}

\begin{thebibliography}{54}
\providecommand{\natexlab}[1]{#1}
\providecommand{\url}[1]{\texttt{#1}}
\expandafter\ifx\csname urlstyle\endcsname\relax
  \providecommand{\doi}[1]{doi: #1}\else
  \providecommand{\doi}{doi: \begingroup \urlstyle{rm}\Url}\fi

\bibitem[AI@Meta(2024)]{llama3modelcard}
AI@Meta.
\newblock Llama 3 model card.
\newblock 2024.
\newblock URL \url{https://github.com/meta-llama/llama3/blob/main/MODEL_CARD.md}.

\bibitem[Anthropic(2023)]{claude-2}
Anthropic.
\newblock Claude 2, 2023.
\newblock URL \url{https://www.anthropic.com/index/claude-2}.

\bibitem[Askell et~al.(2021)Askell, Bai, Chen, Drain, Ganguli, Henighan, Jones, Joseph, Mann, DasSarma, Elhage, Hatfield-Dodds, Hernandez, Kernion, Ndousse, Olsson, Amodei, Brown, Clark, McCandlish, Olah, and Kaplan]{askell2021general}
Amanda Askell, Yuntao Bai, Anna Chen, Dawn Drain, Deep Ganguli, Tom Henighan, Andy Jones, Nicholas Joseph, Ben Mann, Nova DasSarma, Nelson Elhage, Zac Hatfield-Dodds, Danny Hernandez, Jackson Kernion, Kamal Ndousse, Catherine Olsson, Dario Amodei, Tom Brown, Jack Clark, Sam McCandlish, Chris Olah, and Jared Kaplan.
\newblock A general language assistant as a laboratory for alignment, 2021.

\bibitem[Ben~Abacha \& Demner-Fushman(2019)Ben~Abacha and Demner-Fushman]{MedQsum}
Asma Ben~Abacha and Dina Demner-Fushman.
\newblock On the summarization of consumer health questions.
\newblock In \emph{Proceedings of the 57th Annual Meeting of the Association for Computational Linguistics}, pp.\  2228--2234, Florence, Italy, July 2019. Association for Computational Linguistics.
\newblock \doi{10.18653/v1/P19-1215}.
\newblock URL \url{https://aclanthology.org/P19-1215}.

\bibitem[Brown et~al.(2020)Brown, Mann, Ryder, Subbiah, Kaplan, Dhariwal, Neelakantan, Shyam, Sastry, Askell, Agarwal, Herbert-Voss, Krueger, Henighan, Child, Ramesh, Ziegler, Wu, Winter, Hesse, Chen, Sigler, Litwin, Gray, Chess, Clark, Berner, McCandlish, Radford, Sutskever, and Amodei]{brown2020language}
Tom~B. Brown, Benjamin Mann, Nick Ryder, Melanie Subbiah, Jared Kaplan, Prafulla Dhariwal, Arvind Neelakantan, Pranav Shyam, Girish Sastry, Amanda Askell, Sandhini Agarwal, Ariel Herbert-Voss, Gretchen Krueger, Tom Henighan, Rewon Child, Aditya Ramesh, Daniel~M. Ziegler, Jeffrey Wu, Clemens Winter, Christopher Hesse, Mark Chen, Eric Sigler, Mateusz Litwin, Scott Gray, Benjamin Chess, Jack Clark, Christopher Berner, Sam McCandlish, Alec Radford, Ilya Sutskever, and Dario Amodei.
\newblock Language models are few-shot learners, 2020.

\bibitem[Chen et~al.(2023)Chen, Li, Yan, Wang, Gunaratna, Yadav, Tang, Srinivasan, Zhou, Huang, and Jin]{chen2023alpagasus}
Lichang Chen, Shiyang Li, Jun Yan, Hai Wang, Kalpa Gunaratna, Vikas Yadav, Zheng Tang, Vijay Srinivasan, Tianyi Zhou, Heng Huang, and Hongxia Jin.
\newblock Alpagasus: Training a better alpaca with fewer data, 2023.

\bibitem[Chen et~al.(2024)Chen, Ma, Zhang, Hao, Yan, Nourbakhsh, Yang, McAuley, Petzold, and Wang]{chen2024surveylargelanguagemodels}
Zhiyu~Zoey Chen, Jing Ma, Xinlu Zhang, Nan Hao, An~Yan, Armineh Nourbakhsh, Xianjun Yang, Julian McAuley, Linda Petzold, and William~Yang Wang.
\newblock A survey on large language models for critical societal domains: Finance, healthcare, and law, 2024.
\newblock URL \url{https://arxiv.org/abs/2405.01769}.

\bibitem[Chia et~al.(2023)Chia, Hong, Bing, and Poria]{chia2023instructeval}
Yew~Ken Chia, Pengfei Hong, Lidong Bing, and Soujanya Poria.
\newblock Instructeval: Towards holistic evaluation of instruction-tuned large language models, 2023.

\bibitem[Chiang et~al.(2023)Chiang, Li, Lin, Sheng, Wu, Zhang, Zheng, Zhuang, Zhuang, Gonzalez, Stoica, and Xing]{vicuna2023}
Wei-Lin Chiang, Zhuohan Li, Zi~Lin, Ying Sheng, Zhanghao Wu, Hao Zhang, Lianmin Zheng, Siyuan Zhuang, Yonghao Zhuang, Joseph~E. Gonzalez, Ion Stoica, and Eric~P. Xing.
\newblock Vicuna: An open-source chatbot impressing gpt-4 with 90\%* chatgpt quality, March 2023.
\newblock URL \url{https://lmsys.org/blog/2023-03-30-vicuna/}.

\bibitem[Chung et~al.(2022)Chung, Hou, Longpre, Zoph, Tay, Fedus, Li, Wang, Dehghani, Brahma, Webson, Gu, Dai, Suzgun, Chen, Chowdhery, Castro-Ros, Pellat, Robinson, Valter, Narang, Mishra, Yu, Zhao, Huang, Dai, Yu, Petrov, Chi, Dean, Devlin, Roberts, Zhou, Le, and Wei]{chung2022scaling}
Hyung~Won Chung, Le~Hou, Shayne Longpre, Barret Zoph, Yi~Tay, William Fedus, Yunxuan Li, Xuezhi Wang, Mostafa Dehghani, Siddhartha Brahma, Albert Webson, Shixiang~Shane Gu, Zhuyun Dai, Mirac Suzgun, Xinyun Chen, Aakanksha Chowdhery, Alex Castro-Ros, Marie Pellat, Kevin Robinson, Dasha Valter, Sharan Narang, Gaurav Mishra, Adams Yu, Vincent Zhao, Yanping Huang, Andrew Dai, Hongkun Yu, Slav Petrov, Ed~H. Chi, Jeff Dean, Jacob Devlin, Adam Roberts, Denny Zhou, Quoc~V. Le, and Jason Wei.
\newblock Scaling instruction-finetuned language models, 2022.

\bibitem[Dua et~al.(2019)Dua, Wang, Dasigi, Stanovsky, Singh, and Gardner]{dua2019dropreadingcomprehensionbenchmark}
Dheeru Dua, Yizhong Wang, Pradeep Dasigi, Gabriel Stanovsky, Sameer Singh, and Matt Gardner.
\newblock Drop: A reading comprehension benchmark requiring discrete reasoning over paragraphs, 2019.
\newblock URL \url{https://arxiv.org/abs/1903.00161}.

\bibitem[Dubois et~al.(2023)Dubois, Li, Taori, Zhang, Gulrajani, Ba, Guestrin, Liang, and Hashimoto]{dubois2023alpacafarm}
Yann Dubois, Xuechen Li, Rohan Taori, Tianyi Zhang, Ishaan Gulrajani, Jimmy Ba, Carlos Guestrin, Percy Liang, and Tatsunori~B. Hashimoto.
\newblock Alpacafarm: A simulation framework for methods that learn from human feedback, 2023.

\bibitem[Gao et~al.(2021)Gao, Tow, Biderman, Black, DiPofi, Foster, Golding, Hsu, McDonell, Muennighoff, Phang, Reynolds, Tang, Thite, Wang, Wang, and Zou]{eval-harness}
Leo Gao, Jonathan Tow, Stella Biderman, Sid Black, Anthony DiPofi, Charles Foster, Laurence Golding, Jeffrey Hsu, Kyle McDonell, Niklas Muennighoff, Jason Phang, Laria Reynolds, Eric Tang, Anish Thite, Ben Wang, Kevin Wang, and Andy Zou.
\newblock A framework for few-shot language model evaluation, September 2021.
\newblock URL \url{https://doi.org/10.5281/zenodo.5371628}.

\bibitem[Geva et~al.(2021)Geva, Khashabi, Segal, Khot, Roth, and Berant]{geva2021strategyqa}
Mor Geva, Daniel Khashabi, Elad Segal, Tushar Khot, Dan Roth, and Jonathan Berant.
\newblock {Did Aristotle Use a Laptop? A Question Answering Benchmark with Implicit Reasoning Strategies}.
\newblock \emph{Transactions of the Association for Computational Linguistics (TACL)}, 2021.

\bibitem[Han et~al.(2023)Han, Adams, Papaioannou, Grundmann, Oberhauser, L{\"o}ser, Truhn, and Bressem]{han2023medalpaca}
Tianyu Han, Lisa~C Adams, Jens-Michalis Papaioannou, Paul Grundmann, Tom Oberhauser, Alexander L{\"o}ser, Daniel Truhn, and Keno~K Bressem.
\newblock Medalpaca--an open-source collection of medical conversational ai models and training data.
\newblock \emph{arXiv preprint arXiv:2304.08247}, 2023.

\bibitem[Hendrycks et~al.(2021)Hendrycks, Burns, Basart, Zou, Mazeika, Song, and Steinhardt]{hendrycks2021measuring}
Dan Hendrycks, Collin Burns, Steven Basart, Andy Zou, Mantas Mazeika, Dawn Song, and Jacob Steinhardt.
\newblock Measuring massive multitask language understanding, 2021.

\bibitem[Hinton et~al.(2015)Hinton, Vinyals, and Dean]{hinton2015distilling}
Geoffrey Hinton, Oriol Vinyals, and Jeff Dean.
\newblock Distilling the knowledge in a neural network, 2015.

\bibitem[Jin et~al.(2021)Jin, Pan, Oufattole, Weng, Fang, and Szolovits]{jin2021disease}
Di~Jin, Eileen Pan, Nassim Oufattole, Wei-Hung Weng, Hanyi Fang, and Peter Szolovits.
\newblock What disease does this patient have? a large-scale open domain question answering dataset from medical exams.
\newblock \emph{Applied Sciences}, 11\penalty0 (14):\penalty0 6421, 2021.

\bibitem[Jin et~al.(2019)Jin, Dhingra, Liu, Cohen, and Lu]{jin2019pubmedqa}
Qiao Jin, Bhuwan Dhingra, Zhengping Liu, William~W. Cohen, and Xinghua Lu.
\newblock Pubmedqa: A dataset for biomedical research question answering, 2019.

\bibitem[Kirkpatrick et~al.(2017)Kirkpatrick, Pascanu, Rabinowitz, Veness, Desjardins, Rusu, Milan, Quan, Ramalho, Grabska-Barwinska, et~al.]{kirkpatrick2017overcoming}
James Kirkpatrick, Razvan Pascanu, Neil Rabinowitz, Joel Veness, Guillaume Desjardins, Andrei~A Rusu, Kieran Milan, John Quan, Tiago Ramalho, Agnieszka Grabska-Barwinska, et~al.
\newblock Overcoming catastrophic forgetting in neural networks.
\newblock \emph{Proceedings of the national academy of sciences}, 114\penalty0 (13):\penalty0 3521--3526, 2017.

\bibitem[Kung et~al.(2023)Kung, Cheatham, Medenilla, Sillos, De~Leon, Elepa{\~n}o, Madriaga, Aggabao, Diaz-Candido, Maningo, et~al.]{kung2023performance}
Tiffany~H Kung, Morgan Cheatham, Arielle Medenilla, Czarina Sillos, Lorie De~Leon, Camille Elepa{\~n}o, Maria Madriaga, Rimel Aggabao, Giezel Diaz-Candido, James Maningo, et~al.
\newblock Performance of chatgpt on usmle: Potential for ai-assisted medical education using large language models.
\newblock \emph{PLoS digital health}, 2\penalty0 (2):\penalty0 e0000198, 2023.

\bibitem[Li et~al.(2022)Li, Chen, Shen, Chen, Zhang, Li, Wang, Qian, Peng, Mao, Chen, and Yan]{li2022explanations}
Shiyang Li, Jianshu Chen, Yelong Shen, Zhiyu Chen, Xinlu Zhang, Zekun Li, Hong Wang, Jing Qian, Baolin Peng, Yi~Mao, Wenhu Chen, and Xifeng Yan.
\newblock Explanations from large language models make small reasoners better, 2022.

\bibitem[Li et~al.(2023{\natexlab{a}})Li, Zhang, Dubois, Taori, Gulrajani, Guestrin, Liang, and Hashimoto]{alpaca_eval}
Xuechen Li, Tianyi Zhang, Yann Dubois, Rohan Taori, Ishaan Gulrajani, Carlos Guestrin, Percy Liang, and Tatsunori~B. Hashimoto.
\newblock Alpacaeval: An automatic evaluator of instruction-following models.
\newblock \url{https://github.com/tatsu-lab/alpaca_eval}, 2023{\natexlab{a}}.

\bibitem[Li et~al.(2023{\natexlab{b}})Li, Li, Zhang, Dan, Jiang, and Zhang]{li2023chatdoctor}
Yunxiang Li, Zihan Li, Kai Zhang, Ruilong Dan, Steve Jiang, and You Zhang.
\newblock Chatdoctor: A medical chat model fine-tuned on a large language model meta-ai (llama) using medical domain knowledge.
\newblock \emph{Cureus}, 15\penalty0 (6), 2023{\natexlab{b}}.

\bibitem[Lin et~al.(2022)Lin, Hilton, and Evans]{lin2022truthfulqa}
Stephanie Lin, Jacob Hilton, and Owain Evans.
\newblock Truthfulqa: Measuring how models mimic human falsehoods, 2022.

\bibitem[Liévin et~al.(2023)Liévin, Hother, and Winther]{liévin2023large}
Valentin Liévin, Christoffer~Egeberg Hother, and Ole Winther.
\newblock Can large language models reason about medical questions?, 2023.

\bibitem[Longpre et~al.(2023)Longpre, Hou, Vu, Webson, Chung, Tay, Zhou, Le, Zoph, Wei, and Roberts]{longpre2023flan}
Shayne Longpre, Le~Hou, Tu~Vu, Albert Webson, Hyung~Won Chung, Yi~Tay, Denny Zhou, Quoc~V. Le, Barret Zoph, Jason Wei, and Adam Roberts.
\newblock The flan collection: Designing data and methods for effective instruction tuning, 2023.

\bibitem[Nori et~al.(2023{\natexlab{a}})Nori, King, McKinney, Carignan, and Horvitz]{nori2023capabilities}
Harsha Nori, Nicholas King, Scott~Mayer McKinney, Dean Carignan, and Eric Horvitz.
\newblock Capabilities of gpt-4 on medical challenge problems, 2023{\natexlab{a}}.

\bibitem[Nori et~al.(2023{\natexlab{b}})Nori, Lee, Zhang, Carignan, Edgar, Fusi, King, Larson, Li, Liu, Luo, McKinney, Ness, Poon, Qin, Usuyama, White, and Horvitz]{nori2023generalist}
Harsha Nori, Yin~Tat Lee, Sheng Zhang, Dean Carignan, Richard Edgar, Nicolo Fusi, Nicholas King, Jonathan Larson, Yuanzhi Li, Weishung Liu, Renqian Luo, Scott~Mayer McKinney, Robert~Osazuwa Ness, Hoifung Poon, Tao Qin, Naoto Usuyama, Chris White, and Eric Horvitz.
\newblock Can generalist foundation models outcompete special-purpose tuning? case study in medicine, 2023{\natexlab{b}}.

\bibitem[OpenAI(2022)]{openai2022chatgpt}
OpenAI.
\newblock Introducing chatgpt, 2022.
\newblock URL \url{https://openai.com/blog/chatgpt}.
\newblock Accessed: 2023-05-11.

\bibitem[OpenAI(2023)]{openai2023gpt4}
OpenAI.
\newblock Gpt-4 technical report, 2023.

\bibitem[Ouyang et~al.(2022)Ouyang, Wu, Jiang, Almeida, Wainwright, Mishkin, Zhang, Agarwal, Slama, Ray, Schulman, Hilton, Kelton, Miller, Simens, Askell, Welinder, Christiano, Leike, and Lowe]{ouyang2022training}
Long Ouyang, Jeff Wu, Xu~Jiang, Diogo Almeida, Carroll~L. Wainwright, Pamela Mishkin, Chong Zhang, Sandhini Agarwal, Katarina Slama, Alex Ray, John Schulman, Jacob Hilton, Fraser Kelton, Luke Miller, Maddie Simens, Amanda Askell, Peter Welinder, Paul Christiano, Jan Leike, and Ryan Lowe.
\newblock Training language models to follow instructions with human feedback, 2022.

\bibitem[Pal et~al.(2022)Pal, Umapathi, and Sankarasubbu]{pal2022medmcqa}
Ankit Pal, Logesh~Kumar Umapathi, and Malaikannan Sankarasubbu.
\newblock Medmcqa : A large-scale multi-subject multi-choice dataset for medical domain question answering, 2022.

\bibitem[Peng et~al.(2023)Peng, Li, He, Galley, and Gao]{peng2023instruction}
Baolin Peng, Chunyuan Li, Pengcheng He, Michel Galley, and Jianfeng Gao.
\newblock Instruction tuning with gpt-4.
\newblock \emph{arXiv preprint arXiv:2304.03277}, 2023.

\bibitem[Sanh et~al.(2022)Sanh, Webson, Raffel, Bach, Sutawika, Alyafeai, Chaffin, Stiegler, Scao, Raja, Dey, Bari, Xu, Thakker, Sharma, Szczechla, Kim, Chhablani, Nayak, Datta, Chang, Jiang, Wang, Manica, Shen, Yong, Pandey, Bawden, Wang, Neeraj, Rozen, Sharma, Santilli, Fevry, Fries, Teehan, Bers, Biderman, Gao, Wolf, and Rush]{sanh2022multitask}
Victor Sanh, Albert Webson, Colin Raffel, Stephen~H. Bach, Lintang Sutawika, Zaid Alyafeai, Antoine Chaffin, Arnaud Stiegler, Teven~Le Scao, Arun Raja, Manan Dey, M~Saiful Bari, Canwen Xu, Urmish Thakker, Shanya~Sharma Sharma, Eliza Szczechla, Taewoon Kim, Gunjan Chhablani, Nihal Nayak, Debajyoti Datta, Jonathan Chang, Mike Tian-Jian Jiang, Han Wang, Matteo Manica, Sheng Shen, Zheng~Xin Yong, Harshit Pandey, Rachel Bawden, Thomas Wang, Trishala Neeraj, Jos Rozen, Abheesht Sharma, Andrea Santilli, Thibault Fevry, Jason~Alan Fries, Ryan Teehan, Tali Bers, Stella Biderman, Leo Gao, Thomas Wolf, and Alexander~M. Rush.
\newblock Multitask prompted training enables zero-shot task generalization, 2022.

\bibitem[Sharegpt(2023)]{sharegpt}
Sharegpt.
\newblock Sharegpt, 2023.
\newblock URL \url{sharegpt.com}.

\bibitem[Singhal et~al.(2022)Singhal, Azizi, Tu, Mahdavi, Wei, Chung, Scales, Tanwani, Cole-Lewis, Pfohl, et~al.]{singhal2022large}
Karan Singhal, Shekoofeh Azizi, Tao Tu, S~Sara Mahdavi, Jason Wei, Hyung~Won Chung, Nathan Scales, Ajay Tanwani, Heather Cole-Lewis, Stephen Pfohl, et~al.
\newblock Large language models encode clinical knowledge.
\newblock \emph{arXiv preprint arXiv:2212.13138}, 2022.

\bibitem[Singhal et~al.(2023)Singhal, Tu, Gottweis, Sayres, Wulczyn, Hou, Clark, Pfohl, Cole-Lewis, Neal, Schaekermann, Wang, Amin, Lachgar, Mansfield, Prakash, Green, Dominowska, y~Arcas, Tomasev, Liu, Wong, Semturs, Mahdavi, Barral, Webster, Corrado, Matias, Azizi, Karthikesalingam, and Natarajan]{singhal2023expertlevel}
Karan Singhal, Tao Tu, Juraj Gottweis, Rory Sayres, Ellery Wulczyn, Le~Hou, Kevin Clark, Stephen Pfohl, Heather Cole-Lewis, Darlene Neal, Mike Schaekermann, Amy Wang, Mohamed Amin, Sami Lachgar, Philip Mansfield, Sushant Prakash, Bradley Green, Ewa Dominowska, Blaise~Aguera y~Arcas, Nenad Tomasev, Yun Liu, Renee Wong, Christopher Semturs, S.~Sara Mahdavi, Joelle Barral, Dale Webster, Greg~S. Corrado, Yossi Matias, Shekoofeh Azizi, Alan Karthikesalingam, and Vivek Natarajan.
\newblock Towards expert-level medical question answering with large language models, 2023.

\bibitem[Suzgun et~al.(2022)Suzgun, Scales, Schärli, Gehrmann, Tay, Chung, Chowdhery, Le, Chi, Zhou, and Wei]{suzgun2022challenging}
Mirac Suzgun, Nathan Scales, Nathanael Schärli, Sebastian Gehrmann, Yi~Tay, Hyung~Won Chung, Aakanksha Chowdhery, Quoc~V. Le, Ed~H. Chi, Denny Zhou, and Jason Wei.
\newblock Challenging big-bench tasks and whether chain-of-thought can solve them, 2022.

\bibitem[Taori et~al.(2023)Taori, Gulrajani, Zhang, Dubois, Li, Guestrin, Liang, and Hashimoto]{alpaca}
Rohan Taori, Ishaan Gulrajani, Tianyi Zhang, Yann Dubois, Xuechen Li, Carlos Guestrin, Percy Liang, and Tatsunori~B. Hashimoto.
\newblock Stanford alpaca: An instruction-following llama model.
\newblock \url{https://github.com/tatsu-lab/stanford_alpaca}, 2023.

\bibitem[Touvron et~al.(2023{\natexlab{a}})Touvron, Lavril, Izacard, Martinet, Lachaux, Lacroix, Rozière, Goyal, Hambro, Azhar, Rodriguez, Joulin, Grave, and Lample]{touvron2023llama}
Hugo Touvron, Thibaut Lavril, Gautier Izacard, Xavier Martinet, Marie-Anne Lachaux, Timothée Lacroix, Baptiste Rozière, Naman Goyal, Eric Hambro, Faisal Azhar, Aurelien Rodriguez, Armand Joulin, Edouard Grave, and Guillaume Lample.
\newblock Llama: Open and efficient foundation language models, 2023{\natexlab{a}}.

\bibitem[Touvron et~al.(2023{\natexlab{b}})Touvron, Martin, Stone, Albert, Almahairi, Babaei, Bashlykov, Batra, Bhargava, Bhosale, Bikel, Blecher, Ferrer, Chen, Cucurull, Esiobu, Fernandes, Fu, Fu, Fuller, Gao, Goswami, Goyal, Hartshorn, Hosseini, Hou, Inan, Kardas, Kerkez, Khabsa, Kloumann, Korenev, Koura, Lachaux, Lavril, Lee, Liskovich, Lu, Mao, Martinet, Mihaylov, Mishra, Molybog, Nie, Poulton, Reizenstein, Rungta, Saladi, Schelten, Silva, Smith, Subramanian, Tan, Tang, Taylor, Williams, Kuan, Xu, Yan, Zarov, Zhang, Fan, Kambadur, Narang, Rodriguez, Stojnic, Edunov, and Scialom]{touvron2023llama-2}
Hugo Touvron, Louis Martin, Kevin Stone, Peter Albert, Amjad Almahairi, Yasmine Babaei, Nikolay Bashlykov, Soumya Batra, Prajjwal Bhargava, Shruti Bhosale, Dan Bikel, Lukas Blecher, Cristian~Canton Ferrer, Moya Chen, Guillem Cucurull, David Esiobu, Jude Fernandes, Jeremy Fu, Wenyin Fu, Brian Fuller, Cynthia Gao, Vedanuj Goswami, Naman Goyal, Anthony Hartshorn, Saghar Hosseini, Rui Hou, Hakan Inan, Marcin Kardas, Viktor Kerkez, Madian Khabsa, Isabel Kloumann, Artem Korenev, Punit~Singh Koura, Marie-Anne Lachaux, Thibaut Lavril, Jenya Lee, Diana Liskovich, Yinghai Lu, Yuning Mao, Xavier Martinet, Todor Mihaylov, Pushkar Mishra, Igor Molybog, Yixin Nie, Andrew Poulton, Jeremy Reizenstein, Rashi Rungta, Kalyan Saladi, Alan Schelten, Ruan Silva, Eric~Michael Smith, Ranjan Subramanian, Xiaoqing~Ellen Tan, Binh Tang, Ross Taylor, Adina Williams, Jian~Xiang Kuan, Puxin Xu, Zheng Yan, Iliyan Zarov, Yuchen Zhang, Angela Fan, Melanie Kambadur, Sharan Narang, Aurelien Rodriguez, Robert Stojnic, Sergey Edunov, and Thomas
  Scialom.
\newblock Llama 2: Open foundation and fine-tuned chat models, 2023{\natexlab{b}}.

\bibitem[Tran et~al.(2023)Tran, Yang, Yao, and Yu]{Tran2023BioInstructIT}
Hieu Tran, Zhichao Yang, Zonghai Yao, and Hong Yu.
\newblock Bioinstruct: Instruction tuning of large language models for biomedical natural language processing.
\newblock \emph{ArXiv}, abs/2310.19975, 2023.
\newblock URL \url{https://api.semanticscholar.org/CorpusID:264744285}.

\bibitem[Vilares \& G{\'o}mez-Rodr{\'i}guez(2019)Vilares and G{\'o}mez-Rodr{\'i}guez]{headqa}
David Vilares and Carlos G{\'o}mez-Rodr{\'i}guez.
\newblock {HEAD}-{QA}: A healthcare dataset for complex reasoning.
\newblock In \emph{Proceedings of the 57th Annual Meeting of the Association for Computational Linguistics}, pp.\  960--966, Florence, Italy, July 2019. Association for Computational Linguistics.
\newblock \doi{10.18653/v1/P19-1092}.
\newblock URL \url{https://www.aclweb.org/anthology/P19-1092}.

\bibitem[Wang et~al.(2023{\natexlab{a}})Wang, Li, Chen, Cai, Zhu, Lin, Cao, Liu, Liu, and Sui]{wang2023large}
Peiyi Wang, Lei Li, Liang Chen, Zefan Cai, Dawei Zhu, Binghuai Lin, Yunbo Cao, Qi~Liu, Tianyu Liu, and Zhifang Sui.
\newblock Large language models are not fair evaluators, 2023{\natexlab{a}}.

\bibitem[Wang et~al.(2023{\natexlab{b}})Wang, Kordi, Mishra, Liu, Smith, Khashabi, and Hajishirzi]{wang2023selfinstruct}
Yizhong Wang, Yeganeh Kordi, Swaroop Mishra, Alisa Liu, Noah~A. Smith, Daniel Khashabi, and Hannaneh Hajishirzi.
\newblock Self-instruct: Aligning language models with self-generated instructions, 2023{\natexlab{b}}.

\bibitem[Wei et~al.(2022)Wei, Bosma, Zhao, Guu, Yu, Lester, Du, Dai, and Le]{wei2022finetuned}
Jason Wei, Maarten Bosma, Vincent~Y. Zhao, Kelvin Guu, Adams~Wei Yu, Brian Lester, Nan Du, Andrew~M. Dai, and Quoc~V. Le.
\newblock Finetuned language models are zero-shot learners, 2022.

\bibitem[Wu et~al.(2023)Wu, Lin, Zhang, Zhang, Wang, and Xie]{wu2023pmcllama}
Chaoyi Wu, Weixiong Lin, Xiaoman Zhang, Ya~Zhang, Yanfeng Wang, and Weidi Xie.
\newblock Pmc-llama: Towards building open-source language models for medicine, 2023.

\bibitem[Xie et~al.(2024)Xie, Chen, Chen, Peng, Hu, Lin, Peng, Huang, Zhang, Keloth, Zhou, He, Ohno-Machido, Wu, Xu, and Bian]{Xie2024MeLF}
Qianqian Xie, Qingyu Chen, Aokun Chen, C.A.I. Peng, Yan Hu, Fongci Lin, Xueqing Peng, Jimin Huang, Jeffrey Zhang, Vipina~Kuttichi Keloth, Xingyu Zhou, Huan He, Lucila Ohno-Machido, Yonghui Wu, Hua Xu, and Jiang Bian.
\newblock Me llama: Foundation large language models for medical applications.
\newblock \emph{ArXiv}, abs/2402.12749, 2024.
\newblock URL \url{https://api.semanticscholar.org/CorpusID:267759846}.

\bibitem[Xu et~al.(2023{\natexlab{a}})Xu, Sun, Zheng, Geng, Zhao, Feng, Tao, and Jiang]{xu2023wizardlm}
Can Xu, Qingfeng Sun, Kai Zheng, Xiubo Geng, Pu~Zhao, Jiazhan Feng, Chongyang Tao, and Daxin Jiang.
\newblock Wizardlm: Empowering large language models to follow complex instructions, 2023{\natexlab{a}}.

\bibitem[Xu et~al.(2023{\natexlab{b}})Xu, Guo, Duan, and McAuley]{xu2023baize}
Canwen Xu, Daya Guo, Nan Duan, and Julian McAuley.
\newblock Baize: An open-source chat model with parameter-efficient tuning on self-chat data, 2023{\natexlab{b}}.

\bibitem[Zhang et~al.(2023)Zhang, Li, Yang, Tian, Qin, and Petzold]{zhang2023enhancing}
Xinlu Zhang, Shiyang Li, Xianjun Yang, Chenxin Tian, Yao Qin, and Linda~Ruth Petzold.
\newblock Enhancing small medical learners with privacy-preserving contextual prompting, 2023.

\bibitem[Zheng et~al.(2023)Zheng, Chiang, Sheng, Zhuang, Wu, Zhuang, Lin, Li, Li, Xing, Zhang, Gonzalez, and Stoica]{zheng2023judging}
Lianmin Zheng, Wei-Lin Chiang, Ying Sheng, Siyuan Zhuang, Zhanghao Wu, Yonghao Zhuang, Zi~Lin, Zhuohan Li, Dacheng Li, Eric.~P Xing, Hao Zhang, Joseph~E. Gonzalez, and Ion Stoica.
\newblock Judging llm-as-a-judge with mt-bench and chatbot arena, 2023.

\bibitem[Zhou et~al.(2023)Zhou, Liu, Xu, Iyer, Sun, Mao, Ma, Efrat, Yu, Yu, Zhang, Ghosh, Lewis, Zettlemoyer, and Levy]{zhou2023lima}
Chunting Zhou, Pengfei Liu, Puxin Xu, Srini Iyer, Jiao Sun, Yuning Mao, Xuezhe Ma, Avia Efrat, Ping Yu, Lili Yu, Susan Zhang, Gargi Ghosh, Mike Lewis, Luke Zettlemoyer, and Omer Levy.
\newblock Lima: Less is more for alignment, 2023.

\end{thebibliography}
\bibliographystyle{iclr2025_conference}

\clearpage
\newpage

\appendix

\clearpage
\newpage
\section*{Limitations}
Our approach utilizes `teacher' LLMs, such as GPT-4 and ChatGPT, to automatically generate medical instruction-response pair datasets, employing these teacher models as medical knowledge bases. However, this could result in hallucinations in the medical knowledge generation. To enhance the generation reliability, we aim to integrate LLMs with the internet and knowledge graphs in future work.

\section*{Ethics Statement}

This research demonstrates the potential of enhancing open-source LMs’ medical capacity and general applicability by distilling knowledge from more powerful "teacher" LLMs. We improve the smaller models’ ability to follow medical instructions and align with user intentions, offering potential benefits for future healthcare applications. Even though our model \textit{AlpaCare} maintains comparable truthfulness scores compared to other baselines, but the rate of correct answers is
still low, showing that our model is likely to hallucinate incorrect answers. Therefore, it is crucial to emphasize that this system is designed to serve as an assistant tool, complementing but not replacing the expertise and judgment of healthcare professionals. All outputs generated by this system must be rigorously validated by licensed medical practitioners before any practical application.

The transition of these LMs to practical medical scenarios — useful for healthcare professionals, patients, and other medical personnel — necessitates extensive research to ensure safety, privacy, and reliability. This development involves not only technical robustness but also a commitment to ethical standards. It includes comprehensive quality assessments in various clinical environments to ensure that the system meets the highest standards of accuracy and ethical conduct. Additionally, attention must be paid to the potential implications for patient privacy and data security, ensuring that all patient information is handled with the utmost confidentiality and in compliance with relevant data protection laws and healthcare regulations.

\section{Data and Code} \label{app: data}
We provide codebase, \textit{MedInstruct-52k} and \textit{MedInstruct-test}, and AlpaCare models in the link \url{https://anonymous.4open.science/r/AlpaCare-D6BB/}.

\section{Medical Task Difficulty Level Scoring System}\label{app:difficulty}

We introduce a clinician-crafted seed set to generate \textit{MedInstruct-52k} and a free-form medical instruction evaluation set, \textit{MedInstruct-test}. This set spans a medical difficulty scale ranging from 1 to 5, where 1 represents the easiest tasks and 5 indicates the most challenging ones. A clinician assessed the difficulty levels of all instances within both the seed set and \textit{MedInstruct-test} based on the scoring system shown in Table \ref{tab:score_system} This system offers a refined dimension for prompting GPT-4 to produce tasks across varied difficulty levels and to evaluate medical proficiency of instruction-tuned models.

\begin{table}[t!]
    \centering
            \caption{Scoring system for evaluating the difficulty level of medical tasks.}

    \begin{tabular}{c|p{7cm}}
    \toprule
         Score & Description\\
         \hline
        1 &  The fact is very basic. The answer becomes apparent immediately after reading the question, or it can be easily found through a direct internet search.\\
        & \\
        2 & The fact is simple but may require a slight application of real-world knowledge, rephrasing, or extending the information to find the answer.\\
               & \\
        3 &This involves facts that require more real-world application, dealing with practical and somewhat complicated situations. It may require more complex paraphrasing and/or communication skills, such as emotional support, psychological evaluations, and ethical considerations. The tested knowledge in this category can be quite challenging.\\
               & \\
        4 & This level involves complicated medical facts. Answering questions at this level may require multi-step thinking processes. The questions might be lengthy and detailed, necessitating simplification for a clearer answer. This category might include most USMLE questions. It may also require a demonstration of enhanced emotional support, psychological evaluations, and ethical considerations. Questions might be based on vague symptom descriptions, making the diagnosis challenging, or involve recent advancements, publications, or current global health issues like pandemics.\\
               & \\
        5 & This category involves complex medical knowledge applied to real-world, intricate situations. The questions are detailed and lengthy, often requiring simplification and multi-step thinking to answer. Some questions might be based on actual medical cases with challenging diagnoses and treatments. The symptom descriptions might be highly vague. Questions could also involve new technologies, recent publications, or current pandemics, requiring decision-making or choosing the best available option. Instructions might also necessitate the demonstration of humane care.\\
         \bottomrule
    \end{tabular}
    \label{tab:score_system}
\end{table}

\clearpage
\newpage

\section{Prompt details for \textit{ MedInstruct-52k} generation} \label{app:generation_prompt}
Here we provide prompts we use for query GPT-4 and ChatGPT for task and response generation.
\begin{table}[h!]
    \centering
        \caption{    
   Task generation prompt
    }
    \vspace{-3mm}
    \begin{tabular}{p{8cm}}
        \toprule

Your objective is to generate diverse medical-related tasks.\\
\\
Here are the requirements:\\
1. Ensure that all tasks are related to the medical domain.\\
2. Craft tasks that encompass varied points of view, e.g. experts, students and patients, etc.\\
3. Maximize the range of task topics, e.g. diseases, treatment, diagnoses, epidemiology, pharmacology, pathophysiology, anatomy, genetics, medical education, etc.\\
4. Introduce different task formats, e.g. text generation, open Q\&A, chat, rewrites, summarizations, classifications, USMLE style Q\&A, multiple-choice Q\&A, single-hop reasoning and multiple-hop reasoning etc.\\
5. All the formats specified in point 4 MUST be represented in the task you generate.\\
6. Create tasks with medical difficulty levels from 1 to 5, with 1 being the easiest and 5 the hardest.\\
7. Use diverse language in the instructions. For instance, combine questions with imperative forms.\\
8. Some instructions might require specific inputs. If an input is not necessary, such as with general instructions like "What are the side effects of COVID-19?", use "<noinput>" in the input field.\\
9. When provided, inputs must range between 50 to 200 words and offer detailed medical context , e.g. symptom descriptions, radiology reports, clinical notes, and exam questions, etc. \\
10. Generate a detailed and comprehensive input instead ask user-provided input.\\
11. Ensure USMLE style Q\&A and multiple-choice Q\&A tasks have both question and choices in input, and the question context should be detailed.\\
12. The USMLE-style question length must exceed 50 words.\\
13. Match instruction and input to the task's perspective. Patient perspectives should be simple and in first person, while clinician views should have professional terminology.\\
14. Ensure the lengths of inputs for different tasks are notably distinct.\\
15. Each task should adhere to the following structure: 'Type: \textbackslash n, Topic: \textbackslash n, View: \textbackslash n, Difficulty: \textbackslash n, Instruction: \textbackslash n, Input: '. Start each new task with '\#\#\#'.\\
\\
List of  15 tasks: \\

Seed task 1 \\
Seed Task 2 \\
Seed Task 3\\

    \bottomrule
    \end{tabular}
    
\end{table}

\begin{table}[h!]
    \centering
        \caption{    
   Output generation prompt
    }
    \begin{tabular}{p{8cm}}
        \toprule

You are a medical expert tasked with answering various medical questions. You MUST generate your response based on the requirements.\\
\\
Here are the requirements:\\
1. For multiple-choice, calculation, and classification problems, you can generate intermediate thinking steps if necessary; otherwise, provide the final answer directly.\\
2. All the intermediate thinking steps must be generated before final answer.\\
3. For multiple-choice questions, you MUST generate the answer choice in the following format: `The answer is (your choice).' For example:\\
`Choose the correct answer. Where in your body will you find the tibia bone? A) Arm B) Foot C) Skull D) Leg\\ 
The tibia bone is one of the two bones in the lower leg, the other being the fibula. The answer is D) Leg.'\\
4. For other types of questions, except multiple-choice, do not use the format mentioned in point 3.\\
\\
task instruction \\
task input (if exist)\\
    \bottomrule
    \end{tabular}
\end{table}

\newpage 
\section{LLM APIs cost analysis and selection reasons} \label{app:cost}
We spent approximately \$900 on task generation using GPT-4 and \$500 on response generation using GPT-3.5-Turbo. The cost of using the GPT-4 API is 30 times higher than that of GPT-3.5-Turbo at the time we created the dataset, which would increase the response cost to \$15,000, making it difficult to stay within budget. Therefore, we utilized GPT-3.5-Turbo as the response generator and one of our evaluators to balance cost and effectiveness.
To demonstrate the suitability of using GPT-3.5-Turbo for dataset generation, we verified the quality of MedInstruct-52k by randomly selecting 50 instances and having a clinician evaluate the responses generated by GPT-3.5-Turbo. The clinician found 49 out of 50 responses to be correct, demonstrating the dataset's high quality.

\section{Training hyperparameter details} 
\label{app:hyper}
Here we provide hyperparameter details for AlpaCare training.
\begin{table}[h!]
\centering
  \caption{
   \textbf{\textit{AlpaCare} hyperparameter setup.} 
  }
  \resizebox{\linewidth}{!}{
    \begin{tabular}{cccccc}
     \toprule
    Model Size &  Data Size & GPUs  & Epoch & LR& Batch Size\\
    \midrule
    7B & 52k& 4 40G A100 & 3 & 2e-5 & 128\\
    13B & 52k& 4 80G A100 & 5 & 1e-5 & 128\\
     \bottomrule  
    \end{tabular}
  }
\label{tab:judge_type_more}
\end{table}

\section{\textit{MedInstruct-test} statistics}   \label{app:test_stat}
Here, we present the test dataset statistics in Table \ref{table:test_stat}.
\begin{table}[h!]
  \centering
\caption{\textbf{\textit{MedInstruct-test} statistics.} The distribution of task counts across various difficulty levels in \textit{MedInstruct-test} is approximately equal to comprehensively evaluate medical proficiency.}
  \resizebox{0.65\linewidth}{!}{
    \begin{tabular}{lccccc}
      \toprule
      Difficulty Level & 1 & 2 & 3 & 4 & 5 \\
      \midrule
      Count & 44 & 46 & 41 & 41 & 44 \\
      \bottomrule  
    \end{tabular}
  }
  \label{table:test_stat}
\end{table}

\section{More experimental results} \label{app:moreexp}

\subsection{General domain free-form instruction evaluation}

\begin{table}[h!]
\centering
  \caption{\textbf{Comparison on general domain free-form instruction evaluation.} A performance comparison between \textit{AlpaCare} and baselines on AlpacaFarm on 4 distinct reference models: Text-davinci-003, GPT-3.5-turbo, GPT-4 and Claude-2. `AVG' represents the mean performance score across all referenced models.}
    \vspace{-3mm}
  \resizebox{\linewidth}{!}{
    \begin{tabular}{l ccccc }
     \toprule
     & \multicolumn{5}{c}{\textbf{AlpacaFarm}} \\
     \cmidrule(lr){2-6}
     & Text-davinci-003 & GPT-3.5-turbo & GPT-4& Claude-2 & AVG \\
  \midrule
Alpaca & 38.7&20.6&14.5&16.9 &  22.7 \\
ChatDoctor &37.4&20.3&13.1&14.0 & 21.2 \\
Medalpaca & 38.2&24.4&20.6&20.1 & 25.8 \\
PMC &15.8&2.6	&13.3&1.6 & 8.3 \\
Baize-H &29.9&	16.9&12.7&13.7 & 18.3 \\
AlpaCare & \textbf{56.4}& \textbf{38.6}& \textbf{34.2}& \textbf{33.7} & \textbf{40.7} \\
     \bottomrule  
    \end{tabular}
  }
 \label{tab:gen-free-form_all}
\end{table}

\subsection{More analysis in general domain performance}

\begin{table}[h!]
    \centering
    \begin{tabular}{lcc}
        \toprule
       Model & StrategyQA& DROP \\
        \midrule
        Alpaca & 57.80 & 23.68 \\
        AlpaCare & 58.02 & 24.96 \\
         \bottomrule  
    \end{tabular}
    \caption{Performance of Alpaca and AlpaCare on StrategyQA and DROP datasets.}
        \vspace{-3mm}
    \label{app:moregeneral}
\end{table}

Compared to Alpaca \citep{alpaca}, AlpaCare achieves better results in the general domain. This improvement is likely due to the intensive knowledge and reasoning embedded in the medical dataset \citep{liévin2023large,chen2024surveylargelanguagemodels}. For example, in the BBH results, the top three categories where AlpaCare outperforms Alpaca are `dyck\_languages', `movie\_recommendation', and `navigate', which requires strong knowledge and reasoning abilities. To further support these findings, we conducted a knowledge-intensive commonsense evaluation using StrategyQA \citep{geva2021strategyqa} and an additional reasoning benchmark evaluation using DROP \citep{dua2019dropreadingcomprehensionbenchmark} to compare Alpaca and AlpaCare with LLaMA \citep{touvron2023llama}-7B as backbone, following the methodologies in \citet{eval-harness} and \citet{chia2023instructeval}, respectively. The Table \ref{app:moregeneral}  presents the results.

These results reinforce that AlpaCare's enhanced performance is not limited to the medical domain but also extends to broader general domain tasks, thereby confirming its superior generalizability.

\subsection{Ablation study} \label{app:abstudy}
We show the detailed score of 4 reference models for medical free-form instruction evaluation on 13B instruction-tuned models in Table \ref{tab:13_b_more}.
\begin{table}[h!]
\centering
  \caption{
   \textbf{Result comparison of 4 reference models on 13B instruction-tuned models.} 
  }
  \resizebox{0.75\linewidth}{!}{
    \begin{tabular}{l ccccc}
     \toprule
     & \multicolumn{5}{c}{\textbf{iCliniq}} \\
     & Text-davinci-003 & GPT-3.5-turbo & GPT-4& Claude-2 & AVG\\
      \cmidrule(lr){2-6}
Alpaca & 46.7&37.0 &19.6&21.7 &  31.3 \\
Medalpaca & 8.1&4.4&1.0 &2.0 & 3.9 \\
PMC &40.6&29.0&14.3&17.5 &  25.4\\
AlpaCare &\textbf{66.7}& \textbf{51.2}& \textbf{48.2}& \textbf{50.2}  & \textbf{54.4} \\
  \midrule
     & \multicolumn{5}{c}{\textbf{MedInstruct}} \\
     & Text-davinci-003 & GPT-3.5-turbo & GPT-4& Claude-2 & AVG \\
      \cmidrule(lr){2-6}
Alpaca & 39.8&22.5&27.1&18.1 &  26.9 \\
Medalpaca & 0.2&0&0&0& 0.1 \\
PMC &44.9&31.9&32.8&29.2 & 34.7\\
AlpaCare & \textbf{71.3}& \textbf{49.1}& \textbf{49.8}& \textbf{47.7} & \textbf{54.5} \\
     \bottomrule  
    \end{tabular}
  }
\label{tab:13_b_more}
\end{table}

We show the detailed score of 4 reference models for medical free-form instruction evaluation on different backbones in Table \ref{tab:model_type_more}.
\begin{table}[h!]
\centering
     \Huge
  \caption{\textbf{Results on different LLM backbone across 4 reference models by using gpt-3.5-tubro as the judge.} Comparing the performance of AlpaCare and Alpaca using different LLM backbones, with 4 distinct reference models.}
      \vspace{-3mm}
  \resizebox{0.75\linewidth}{!}{%
	\begin{tabular}{l l ccccc}
	 \toprule
     & &\multicolumn{5}{c}{\textbf{iCliniq}} \\ 
     & & Text-davinci-003 & GPT-3.5-turbo & GPT-4& Claude-2 & AVG\\
      \cmidrule(lr){3-7}  
	 \multirow{2}{*}{LLaMA} & Alpaca & 38.8 & 30.4 & 12.8 & 15.6 & 24.4 \\
	 & AlpaCare & \textbf{66.6} & \textbf{50.6} & \textbf{47.4} & \textbf{49.7} &\textbf{53.6} \\ 
  	 \multirow{2}{*}{LLaMA-2} & Alpaca & 45.8& 36.3& 18.2& 20.8& 30.3 \\
	 & AlpaCare & \textbf{66.5}& \textbf{50.4}& \textbf{47.8} & \textbf{50}&  \textbf{53.7} \\ 
    	 \multirow{2}{*}{LLaMA-3} & Alpaca & 42.3& 28.6& 26.4&10.0& 26.8\\
	 & AlpaCare & \textbf{77.6}& \textbf{53.9}& \textbf{46.3} & \textbf{49.7}&   \textbf{56.9} \\
  	\bottomrule
  & &\multicolumn{5}{c}{\textbf{MedInstruct}} \\
    & & Text-davinci-003 & GPT-3.5-turbo & GPT-4& Claude-2 & AVG\\
      \cmidrule(lr){3-7}
	 \multirow{2}{*}{LLaMA} & Alpaca & 35.0 & 20.6 & 21.5 & 15.6 & 24.4  \\
	 & AlpaCare &  \textbf{67.6} & \textbf{48.8} & \textbf{47.4} & \textbf{49.7} &\textbf{53.5}   \\
  	 \multirow{2}{*}{LLaMA-2} & Alpaca & 39.6& 22.7& 26.4&18.5& 26.8\\
	 & AlpaCare & \textbf{70.6}& \textbf{48.8}& \textbf{50.0} & \textbf{48.4}&   \textbf{54.2}  \\
  \multirow{2}{*}{LLaMA-3} & Alpaca & 38.4& 16.9& 14.6&13.0& 20.7\\
	 & AlpaCare & \textbf{78.5}& \textbf{50.0}& \textbf{51.4} & \textbf{46.5}&   \textbf{56.6} 
\\
	 \bottomrule
	\end{tabular}
  }
 \label{tab:model_type_more}
\end{table}

We show the detailed score of 4 reference models for medical free-form instruction evaluation by using Claude-2 as the judge in Table \ref{tab:judge_type_more}.
\begin{table}[t!]
\centering
  \caption{
   \textbf{Medical free-from instruction evaluation results by using Claude-2 as judge.} 
  }
  \resizebox{0.75\linewidth}{!}{
    \begin{tabular}{l ccccc}
     \toprule
     & \multicolumn{5}{c}{\textbf{iCliniq}} \\
     & Text-davinci-003 & GPT-3.5-turbo & GPT-4& Claude-2 & AVG\\
      \cmidrule(lr){2-6}
Alpaca & 40.7&33.4&18.5&14.2 &  26.7 \\
ChatDoctor &24.5&23.8&11.3&9.9 & 17.4 \\
Medalpaca & 38.3&37.1&15.8&15.4 & 26.7 \\
PMC &0&0	&1.9&3.1 & 1.3 \\
Baize-H &46.9&	35.6&12.1&7.5 & 25.5 \\
AlpaCare &\textbf{64.5}& \textbf{46.8}& \textbf{26.9}& \textbf{17.1}  & \textbf{38.8} \\
  \midrule
     & \multicolumn{5}{c}{\textbf{MedInstruct}} \\
     & Text-davinci-003 & GPT-3.5-turbo & GPT-4& Claude-2 & AVG \\
      \cmidrule(lr){2-6}
Alpaca & 44.4&19.9&19.2&13.4 &  23.5 \\
ChatDoctor &34.8&17.5 &14.1 &10.3 & 21.7 \\
Medalpaca & 41.4&19.3&18.0&12.5 & 23.1 \\
PMC &2.8&1.5	&0.1&0.7 & 1.8 \\
Baize-H &40.1&	25.1&21.8&15.2 & 19.8 \\
AlpaCare & \textbf{76.4}& \textbf{42.6}& \textbf{42.9}& \textbf{31.6} & \textbf{31.5} \\
     \bottomrule  
    \end{tabular}
  }
\label{tab:judge_type_more}
\end{table}

\end{document}